\documentclass[journal]{IEEEtran}
\usepackage{amsmath,amsfonts}
\usepackage{algorithm}
\usepackage{algpseudocode}
\usepackage{color}
\usepackage{soul}
\usepackage{cite}
\usepackage{graphicx}
\usepackage{amsthm}
\usepackage{multirow}
\def\BibTeX{{\rm B\kern-.05em{\sc i\kern-.025em b}\kern-.08em
    T\kern-.1667em\lower.7ex\hbox{E}\kern-.125emX}}
\usepackage{subfigure}
\begin{document}
\title{A Universal Cooperative Decision-Making Framework for Connected Autonomous Vehicles with Generic Road Topologies}
\author{Zhenmin Huang, Shaojie Shen, and Jun Ma
%% Initial Version
\thanks{Zhenmin Huang, Shaojie Shen, and Jun Ma are with the Department of Electronic and Computer Engineering, The Hong Kong University of Science and Technology, Hong Kong SAR, China (e-mail: zhuangdf@connect.ust.hk; eeshaojie@ust.hk; jun.ma@ust.hk).

This work has been submitted to the IEEE for possible publication. Copyright may be transferred without notice,	after which this version may no longer be accessible.}}

\markboth{}{}
\maketitle

\begin{abstract} 
%% Cooperative decision-making of connected autonomous vehicles (CAVs) is a long-standing problem due to its non-convexity, nonlinearity, discrete nature, and various road topologies involved in real traffic scenarios. Most of the existing methods are only applicable to one specific scenario with scenario-specified assumptions and therefore their applications in real-world situations are hindered by the Non-enumerable nature of traffic scenarios. In this paper, with the intuition that the topologies of various traffic scenarios can generally be expressed as directed acyclic graphs (DAGs), we proposed a general optimization approach that can potentially solve cooperative decision-making problems corresponding to traffic scenarios with arbitrary road topologies. In the proposed methods, reference paths and time profiles for all CAVs involved are determined in a fully cooperative manner with proper consideration of velocities, accelerations, conflict resolutions, and total traffic efficiency. Based on the representative DAG of road topologies, the cooperative decision-making of CAVs is approximately formulated as a mixed-integer linear programming (MILP) problem, which is readily solvable by general commercial solvers with global optimality attainable. Case studies corresponding to three multi-lane traffic scenarios with different topologies, including straightaway, roundabout, and unsignalized intersection, are provided in simulations part to verify the effectiveness of the proposed method.

Cooperative decision-making of Connected Autonomous Vehicles (CAVs) presents a longstanding challenge due to its inherent nonlinearity, non-convexity, and discrete characteristics, compounded by the diverse road topologies encountered in real-world traffic scenarios. The majority of current methodologies are only applicable to a single and specific scenario, predicated on scenario-specific assumptions. Consequently, their application in real-world environments is restricted by the innumerable nature of traffic scenarios. In this study, we propose a unified optimization approach that exhibits the potential to address cooperative decision-making problems related to traffic scenarios with generic road topologies. This development is grounded in the premise that the topologies of various traffic scenarios can be universally represented as Directed Acyclic Graphs (DAGs). Particularly, the reference paths and time profiles for all involved CAVs are determined in a fully cooperative manner, taking into account factors such as velocities, accelerations, conflict resolutions, and overall traffic efficiency. The cooperative decision-making of CAVs is approximated as a mixed-integer linear programming (MILP) problem building on the DAGs of road topologies. This favorably facilitates the use of standard numerical solvers and the global optimality can be attained through the optimization. Case studies corresponding to different multi-lane traffic scenarios featuring diverse topologies are scheduled as the test itineraries, and the efficacy of our proposed methodology is corroborated.
\end{abstract}

\begin{IEEEkeywords}
Autonomous driving, multi-agent systems, connected autonomous vehicles, cooperative decision-making, non-convex optimization, mixed-integer linear programming (MILP).
\end{IEEEkeywords}

\theoremstyle{definition}
\newtheorem{definition}{Definition}
\theoremstyle{definition}
\newtheorem{remark}{Remark}
\theoremstyle{definition}
\newtheorem{assumption}{Assumption}
\theoremstyle{definition}
\newtheorem{lemma}{Lemma}
\graphicspath{ {pictures/} {pictures/S/} {pictures/R/} {pictures/C/} }

\section{Introduction}

The rapid developments of information technology and artificial intelligence prompt the emergency of connected autonomous vehicles (CAVs), which enable autonomous driving in a cooperative manner. Widely recognized as a promising direction within future transportation systems, CAVs are capable of communicating their driving intentions in real-time with other CAVs, road infrastructures, and cloud devices through Vehicle-to-everything (V2X)~\cite{sun2021survey,chehri2020communication}. As a result, swarm intelligence is enabled and important driving decisions can be made cooperatively to enhance safety, traffic efficiency, and passenger comfort. 
These benefits are particularly evident in complex urban traffic scenarios, such as on-ramp mergings, roundabouts, and unsignalized intersections. In these situations, human drivers often exhibit noncooperative driving behaviors and limited rationality, leading to conflicts and sub-optimal driving choices. For this reason, various methods are proposed for solving cooperative trajectory planning problems of CAVs~\cite{10171831,zhang2021semi}. Nonetheless, developing effective algorithms for obtaining optimal cooperative decisions for CAVs remains a long-standing and challenging problem. This challenge arises from the inherent nonlinearity, non-convexity, and discrete characteristics of cooperative decision-making problems, as well as the diverse topologies associated with different traffic scenarios. To address this issue, researchers have made considerable strides, albeit in a case-by-case approach. As a result, numerous methods have been proposed to address the cooperative decision-making problem in specific scenarios such as on-ramp mergings~\cite{rios2016automated,pei2019cooperative}, roundabouts~\cite{hang2021decision,debada2018virtual}, and intersections~\cite{meng2017analysis,wei2021autonomous}. However, it is important to note that these methods often rely on assumptions and discussions that are closely tied to the specific topologies and geometries of the scenarios to be addressed. As a result, their ability to generalize to different situations is significantly weakened. This lack of generalization poses challenges for the real-world application of these methods, particularly considering the non-enumerable nature of traffic scenarios and their inherent deviations from ideal conditions. Therefore, a comprehensive and general solution to the cooperative decision-making problem for CAVs is yet to be achieved.

%% In this paper, we present a general optimization approach for cooperative decision-making of CAVs, which can potentially be applied to different traffic scenarios with arbitrary road topologies. With the intuition that road topologies can be well-represented by directed acyclic graphs (DAGs), the discussion of the proposed method is based on the representative DAGs and thus decoupled from the actual road topologies. The cooperative decision-making problem of CAVs is then viewed as a multi-agent path-searching problem, which is approximately formulated as a mixed-integer linear programming (MILP) problem. Through the optimization of the corresponding MILP problem, the optimal solution can be obtained, which enables CAVs to pass through complicated traffic scenarios without collisions. Overall system performance is optimized considering important performance indices including velocities, accelerations, and traffic efficiency. The main contributions of this paper are listed as follows:

In this paper, we present a unified optimization approach for cooperative decision-making of CAVs, which is potentially applicable to various traffic scenarios with generic road topologies. Leveraging the fact that road topologies can be effectively represented by directed acyclic graphs (DAGs), the discussion of the proposed method is based on the DAGs, thus allowing it to be independent from the actual road topologies. The cooperative decision-making problem of CAVs is then recasted as a multi-agent path-searching problem, which is approximately formulated as a mixed-integer linear programming (MILP) problem. By solving this MILP problem, we can obtain an optimal solution that enables CAVs to navigate complex traffic scenarios without collisions. Our approach optimizes the overall system performance by considering crucial performance indices, including velocities, accelerations, and traffic efficiency; and this comprehensive perspective ensures that our solution not only enhances safety but also maximizes traffic flow and passenger comfort. The main contributions of this paper are listed as follows:

\begin{itemize}
\item[$\bullet$]
% Based upon discussions over generic graph representations of road topologies, a novel cooperative decision-making framework for CAVs is introduced, which is essentially decoupled from actual road structures and geometries of traffic scenarios. As a result, the proposed framework can be generalized to various traffic scenarios without the need for adaptations.
A novel cooperative decision-making framework for CAVs is presented leveraging generic graph representations of road topologies. This framework is essentially decoupled from actual road structures and geometries of traffic scenarios. Consequently, the proposed framework can be generalized to a wide range of traffic scenarios without the need for specific adaptations.
\end{itemize}

\begin{itemize}
\item[$\bullet$]
% With proper consideration of velocities, accelerations, traffic efficiency, and conflict resolutions, an optimization scheme is introduced such that decisions and time profiles of CAVs are jointly optimized to obtain optimal solutions concerning overall group performance.

Through the active combination of the big-M method and slack variables, an optimization scheme is presented, such that joint optimization of decisions and time profiles can be achieved over arbitrarily complex directed graph representations. 

\end{itemize}

\begin{itemize}
\item[$\bullet$]
% The optimization problem is approximately formulated as an MILP. Through the elimination of nonlinearity, the problem is readily solvable and the global optimality is attainable by applying general commercial solvers for mixed-integer programming purposes.
The optimization problem is formulated as an MILP problem through linearization, which favorably facilitates the use
of standard numerical solvers to obtain the global optimal solution in the cooperative decision-making process.
%such that this problem is readily solvable with commercial solvers and the global optimality is attainable. As a result, optimal solutions can be obtained to maximize the overall group performance.
\end{itemize}

\begin{itemize}
\item[$\bullet$]
% Case studies over three different urban traffic scenarios are provided to verify the effectiveness of the proposed method and demonstrate its ability to generalize to a variety of other traffic scenarios.
To validate our proposed method and demonstrate its adaptability and versatility, we provide case studies on three distinct urban traffic scenarios. These studies showcase the effectiveness of this framework to generalize across a variety of traffic situations.
\end{itemize}

\section{Related Works}

Recently, different categories of methods have been presented to tackle the cooperative decision-making problem of CAVs in various urban traffic scenarios. In particular, methods based on multi-agent reinforcement learning (MARL) are currently under intensive investigation. In~\cite{toghi2021cooperative}, a sympathetic cooperative driving paradigm is proposed through MARL to enable the cooperation between CAVs and human drivers in the merging scenario. In~\cite{antonio2022multi,zhou2019development,peng2021connected}, methods based on MARL are presented to improve the overall traffic efficiencies in signalized/unsignalized intersections. The introduction of learning-based methods into the field of cooperative autonomous driving can assist in solving complex cases that involve intricate vehicle interactions, which are typically hard to model through traditional methods. However, learning-based methods often require large amounts of data and lack interpretability. Besides, these methods are usually fragile to the change in road geometries and topologies, and retraining of the models is often required when generalized to other traffic scenarios. These limitations hinder their widespread application in real-world situations.

In additional to learning-based approaches, miscellaneous methods are proposed from the perspective of heuristic rules~\cite{ding2019rule}, optimization and optimal control~\cite{pan2022convex,zhang2021trajectory}, graph and tree searching~\cite{xu2019cooperative1}, etc. In~\cite{xu2019cooperative2}, the cooperative merging of CAVs is formulated as an optimization problem that aims to minimize the travel time of mainline vehicles and maximize the number of merging vehicles. In~\cite{liu2017distributed}, a distributed mechanism is introduced to provide cooperative driving strategies for CAVs in intersection areas. To resolve conflicts and maximize traffic efficiency, a conflict graph is created and resolved to determine the optimal sequence and allocate time slots for each CAV involved. These methods mainly focus on developing highly dedicated strategies for specific traffic scenarios to achieve improvement in traffic efficiency. Another branch of methods that receives increasing attention is based on game theory, owing to its superiority in modeling the interactions between participating agents. These methods generally formulate the cooperative decision-making problem of CAV as a dynamic game and aim at achieving the corresponding Nash equilibrium or Stackelberg equilibrium strategies, where none of the vehicles involved is able to improve its own utility through unilaterally altering its own strategies. In~\cite{lopez2022game,zhang2019game}, a two-player Stackelberg game and a pair-wise normal game are introduced respectively to tackle the cooperative lane-changing problem in a simple two-lane traffic scenario. Representative works in applying game theory to more complicated traffic scenarios include~\cite{hang2022decision,hang2021decision,hang2021cooperative}, where Stackelberg and coalitional game approaches are successfully utilized to develop strategies for cooperative decision-making of CAVs in traffic scenarios like multi-lane on-ramp merging, roundabout, and intersection. However, in these developments, the formulation of cooperative decision-making as a dynamic game involves dedicated discussion over relative positions of CAVs and road topologies in a case-by-case manner, which impairs the generalization ability of the proposed methods. Except for the game-based methods discussed above, pertinent works also investigate the application of potential games~\cite{monderer1996potential} in cooperative autonomous driving, which is a special class of games that presents several appealing properties. In~\cite{liu2022potential}, a cooperative decision-making framework based on potential games is proposed, which aims to provide robust decision-making strategies across diverse traffic scenarios involving both lane-changing and intersection-crossing tasks. 
%% However, the experiments are presented in a case-by-base manner with the topologies of each scenario overly simplified and handled by designing individual costs, strategies, and action space, implying a lack of generality. In general, game-based methods assume selfish nature of traffic participants. Therefore, cooperation is limited, and global optimality that maximizes the social interest of the entire group is seldom attained.
However, the experiments are presented on a case-by-case basis, with the topologies of each scenario being overly simplified and handled by designing individual costs, strategies, and action spaces. Essentially, this implies a lack of generality in their approaches. In general, game-based methods typically presume a selfish nature among traffic participants. Consequently, these methods provide limited cooperation and rarely achieve global optimality that maximizes the social interest of the entire group.

Meanwhile, many of the methods proposed in this area, including several discussed above, are intertwined with mixed-integer programming (MIP) due to the discrete nature of decision-making. As a branch of optimization problems containing discrete variables, MIP presents extraordinarily expressive capability in modeling problems across different fields but is often hard to solve, especially in the presence of nonlinearities. The utilization of MIP in the field of cooperative decision-making of CAVs is usually two-fold: integer variables are introduced for the construction of pertinent collision avoidance constraints or to represent decisions to be made that are in general discrete. In~\cite{burger2018cooperative,esterle2020optimal},  collision avoidance constraints are established by combining integer variables and the big-M method, yielding a generic mixed-integer quadratic programming (MIQP) scheme for solving the cooperative trajectory planning problem. Meanwhile, developing cooperative decision-making strategies typically requires an active combination of the two usages. In~\cite{fabiani2019multi}, MIP is combined with potential games to generate an approximated mixed-integer Nash equilibrium strategy for cooperative decision-making on multi-lane highways. In~\cite{fayazi2018mixed}, an optimal scheduler based on MILP is introduced to control the intersection crossing of autonomous vehicles. 
In~\cite{dollar2021multilane}, MIQP is utilized for lane selection in highway scenarios. Apparently, these methods are all scenario-specific. 
% Among the existing methods based on MIP, the one presented by~\cite{kessler2019cooperative} is by far the most similar to ours. In the proposed method, possible behaviors of each CAV are sampled at each time step to create a trajectory tree, and collision-free optimal trajectories are obtained through cooperative tree searching, which is achieved by solving a corresponding MILP problem. However, the proposed method fails to consider actual road topologies, and the exponential growth of the scale of the MILP problem makes this method unsuitable for cooperative decision-making over a relatively long horizon.
It is worthwhile to mention that, in~\cite{kessler2019cooperative}, the possible behaviors of each CAV are sampled at each time step to create a trajectory tree. Optimal, collision-free trajectories are then obtained through cooperative tree searching, which is achieved by solving a corresponding MILP problem. However, the proposed method fails to consider actual road topologies, and the exponential growth of the scale of the MILP problem makes this method unsuitable for cooperative decision-making over a relatively long horizon.

\section{Cooperative Decision-Making with Mixed-Integer Linear Programming}
In this section, we propose a generic cooperative decision-making framework applicable to various multi-lane traffic scenarios for CAVs. By representing the road structure of each scenario with a directed waypoint graph, the cooperative decision-making problem of CAVs is decoupled from the road topology and viewed as a cooperative path-searching problem. Lane-selection decisions and other decisions are reflected from the paths on the graph. Through MILP formulation, paths and time profiles along the paths are jointly optimized, yielding an optimal solution concerning overall system performance.
\subsection{Waypoint Graph}

We discretize complicated multi-lane traffic scenarios by sampling waypoints along the centerline of each lane and constructing the connectivity graph over these waypoints according to reachability conditions. The resulting waypoint graph is essentially a directed acyclic graph. Remarkably, the construction of such a waypoint graph is essentially decoupled from the problem formulation and the optimization process, and therefore can be performed beforehand. With the introduction of waypoint graphs, we view CAVs as omnidirectional agents moving along the edges of the graph from starting vertices to desired destinations during the decision-making stage. Thus, the cooperative decision-making process for CAVs then turns into a cooperative path-searching problem over directed acyclic graphs.

\begin{figure}[h]
\centering
\includegraphics[scale=0.21]{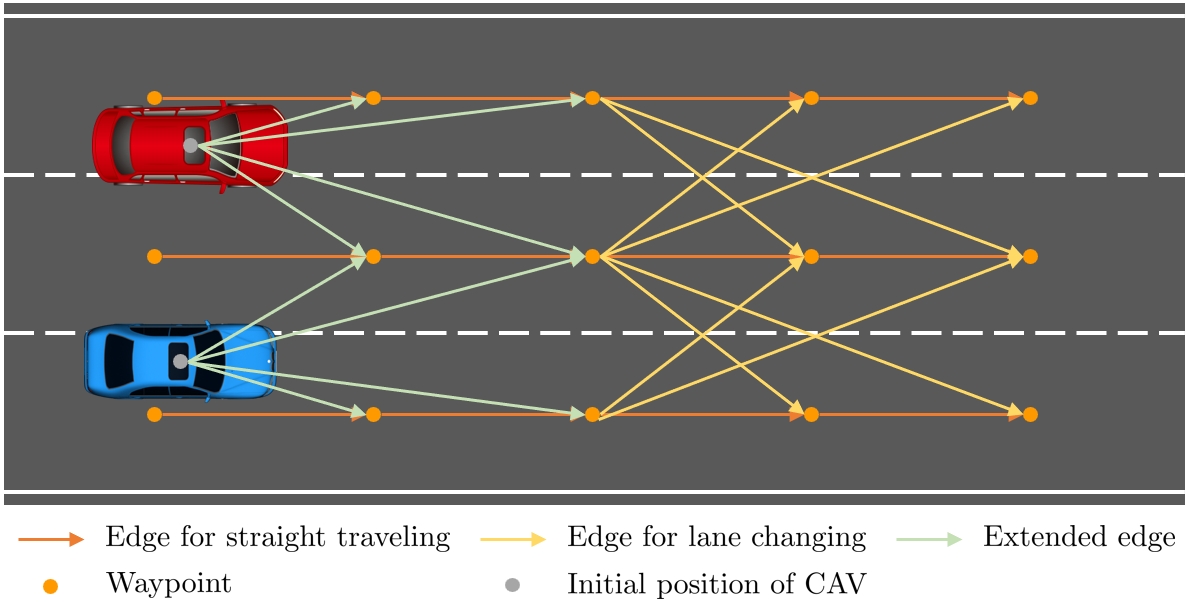}
\caption{An illustrative example of extended waypoint graphs. Different styles of arrows represent edges for different driving purposes, including straight traveling, lane switching, etc. Those edges are treated indiscriminately in the proposed methodology.}
\label{fig:graph}
\end{figure}

An immediate problem that arises from this formulation is that the initial positions of CAVs may not be exactly at any of the vertices in the waypoint graph. Therefore, it is infeasible to identify their starting vertices. This problem can be solved by performing extensions to the original waypoint graph. For each CAV, an extra vertex is added to the graph which is located exactly at its initial position, together with edges that connect this vertex to a fixed number of closest existing vertices in front of the CAV. Such extensions are trivial in general and do not affect the nature of the problem. An illustrative example of extended waypoint graphs is shown in Fig. \ref{fig:graph}.

We denote the extended graph as $\{\mathcal{V},\mathcal{E}\}$, where $\mathcal{V}$ is the set of vertices and $\mathcal{E}$ is the set of directed edges. We also denote the set of CAVs as $\mathcal{N}$. It is then obvious that for a particular CAV $i$ with $i\in\mathcal{N}$, parts of the waypoint graph can be redundant, as CAV $i$ cannot arrive at the vertices that are behind it (reversing is generally not allowed in urban traffic scenarios). To avoid defining redundant variables, we aim to extract a sub-graph $\{\mathcal{V}^i,\mathcal{E}^i\}$ for CAV $i$, such that $\mathcal{V}^i\subset\mathcal{V}$, $\mathcal{E}^i\subset\mathcal{E}$ and $\{\mathcal{V}^i,\mathcal{E}^i\}$ contains vertices and edges relevant to the driving task of the CAV $i$ only. We first present the following definitions:
\begin{definition}
For two vertices $v_1\in\mathcal{V}$ and $v_2\in\mathcal{V}$, we define partial order $v_1\leq v_2$ if there is a directed path from $v_1$ to $v_2$.
\end{definition}
\begin{definition}
For a vertex $v_1\in\mathcal{V}$, we define its forward reachable set $\mathcal{F}(v_1)$ as 
\begin{equation}
    \mathcal{F}(v_1)=\{v_2|v_2\in\mathcal{V},v_1\leq v_2\},
\end{equation}
and its backward reachable set $\mathcal{B}(v_1)$ as
\begin{equation}
    \mathcal{B}(v_1)=\{v_2|v_2\in\mathcal{V},v_2\leq v_1\}.
\end{equation}
\end{definition}

Suppose the starting vertex of CAV $i$ is denoted as $s^i\in\mathcal{V}$, and the set of possible destinations of CAV $i$ is denoted as $des^i\subset\mathcal{V}$. A meaningful vertex for CAV $i$ is the one that is reachable from the starting vertex and reachable to at least one of the destinations. Therefore, $\mathcal{V}^i$ is given as
$\mathcal{V}^i=\mathcal{F}(s^i)\cap(\cup_{v\in des^i}\mathcal{B}(v))$. Meanwhile, $\mathcal{E}^i$ is given as $\mathcal{E}^i=\{e|e\in\mathcal{E},e_1\in\mathcal{V}^i,e_2\in\mathcal{V}^i\}$, where $e_1$ and $e_2$ are the head and tail of $e$, respectively. With the above definitions, we introduce the following important binary decision variables:
\begin{equation}
y^i_e=\left\{
\begin{aligned}
1\ &\textup{if}\ i\ \textup{passes through}\ e \\
0\ &\textup{if}\ i\ \textup{does not pass through}\ e
\end{aligned}
\right.\ \forall i\in\mathcal{N}, \forall e\in\mathcal{E}^i.
\end{equation}

As a result, any path from $s^i$ to $des^i$ can be represented by values of $\{y^i_e\}$. Meanwhile, to obtain the time profile for each CAV, we define $t^i_v$ as the time stamp when the center of vehicle $i$ reaches vertex $v$ for all $i\in\mathcal{N}$ and $v\in\mathcal{V}^i$. Notice that some of $t^i_v$ can still be redundant if the corresponding vehicle $i$ does not pass through the vertex $v$. With the above definitions, the cooperative decision-making problem of CAVs is then turned into the determination of the optimal set of $\{y^i_e,t^i_v\}$ with respect to certain optimization goals subject to relevant constraints.

\subsection{Constraints}
\subsubsection{Path Constraints}
Given that each CAV starts from one of the vertex inside the waypoint graph and heads for its destinations, a valid path will be the one that comes out of the starting vertex and ends in one of the destination vertices with no branches and breaks in between. To enforce the feasibility of the generated paths, we first denote the set of outgoing edges of a particular vertex $v$ as $\mathcal{E}^{out}_v$, and the set of incoming edges of vertex $v$ as $\mathcal{E}^{in}_v$. The following constraints need to be imposed to ensure the feasibility of the generated paths:

\textit{Starting Point Constraints:}
\begin{equation}
\label{start}
\sum_{e\in\mathcal{E}^{i,out}_{s^i}}y^i_e=1,\ \forall i\in\mathcal{N}.
\end{equation}

\textit{Destination Constraints:}
\begin{equation}
\label{destination}
\sum_{v\in des^i}\sum_{e\in\mathcal{E}^{i,in}_{v}}y^i_e=1,\ \forall i\in\mathcal{N}.
\end{equation}

\textit{Continuity Constraints:}
\begin{equation}
\label{continuity}
\sum_{e\in\mathcal{E}^{i,in}_{v}}y^i_e=\sum_{e\in\mathcal{E}^{i,out}_{v}}y^i_e,\ 
\forall i\in\mathcal{N}, \forall v\in\Bar{\mathcal{V}}^i,
\end{equation}
where $\Bar{\mathcal{V}}^i=\{v|v\in\mathcal{V}^i,v\neq s^i, v\notin des^i\}$. Note that these constraints are special cases of flow conservation constraints, where the flows are confined to binary variables, with the starting point and the set of destination points being source and sinks, respectively.

\subsubsection{Velocity Constraints}
In order to improve the comfort of passengers, CAVs are encouraged to pass through the traffic scenarios at roughly steady velocities. Therefore, we set a reference velocity for each of the CAVs to follow. We denote the reference velocity for vehicle $i$ as $V^i_r$. For CAV $i$ that is running along edge $e$, we define the deviation of the average velocity from the reference velocity as $\Delta V^i_e=V^i_e-V^i_r$, where $V^i_e=l_e/(t^i_{e^2}-t^i_{e^i})$ is the average velocity of CAV $i$ along edge $e$. We then propose to minimize $|\Delta V^i_e(t^i_{e^2}-t^i_{e^1})|$, which is the integration of deviation of actual velocity from the reference velocity over time domain $[t^i_{e^1}, t^i_{e^2}]$. In particular, $\Delta V^i_e(t^i_{e^2}-t^i_{e^i})$ is given as
\begin{equation}
    \Delta V^i_e(t^i_{e^2}-t^i_{e^i}) = l_e-V^i_r(t^i_{e^2}-t^i_{e^i}).
\end{equation}
The following constraints are then imposed on the time stamps corresponding to both head and tail of each of the edges:
\begin{equation}
\label{velocity}
\begin{aligned}
&l_e - V^i_r(t^i_{e_2}-t^i_{e_1}) \leq M(1-y^i_e)+\Bar{s}^i_e,\\
&l_e - V^i_r(t^i_{e_2}-t^i_{e_1}) \geq -M(1-y^i_e)-\underline{s}^i_e,\\
&\forall i\in\mathcal{N}, e\in\mathcal{E}^i.
\end{aligned}
\end{equation}
In the above constraints, $e_1$ and $e_2$ are the head vertex and the tail vertex of edge $e$, respectively. $l_e$ is the length of edge $e$. $M$ is a sufficiently large number. $\Bar{s}^i_e$ and $\underline{s}^i_e$ are a pair of positive slack variables that will be penalized in the overall costs. The above constraints are designed such that when $y^i_e=0$, namely the edge is not passed through by vehicle $i$, the constraints are trivially satisfied by arbitrary $t^i_{e_1}$ and $t^i_{e_2}$ given a sufficiently large $M$, and therefore $\Bar{s}^i_e$ and $\underline{s}^i_e$ can be reduced to zero without any consequence. The constraints are thus inactivated without incurring any increase in the overall cost. Meanwhile, when $y^i_e=1$, the constraints are activated, and $|\Delta V^i_e(t^i_{e^2}-t^i_{e^1})|$ is then upper-bounded by $\Bar{s}^i_e+\underline{s}^i_e$. Therefore, penalizing these two slack variables will equally penalize the deviations of velocities.

To ensure that the average velocity $V^i_e$ is bounded, namely $V^i_{slow}\leq V^i_e\leq V^i_{fast}$, $\Bar{s}^i_e$ and $\underline{s}^i_e$ must satisfy the following constraints:
\begin{equation}
\label{velocitySlack}
\begin{aligned}
&0\leq\Bar{s}^i_e\leq(V^i_{fast}-V^i_r)(t^i_{e_2}-t^i_{e_1})+M(1-y^i_e),\\
&0\leq\underline{s}^i_e\leq(V^i_r-V^i_{slow})(t^i_{e_2}-t^i_{e_1})+M(1-y^i_e),\\
&\forall i\in\mathcal{N}, e\in\mathcal{E}^i.
\end{aligned}
\end{equation}

\subsubsection{Longitudinal Acceleration Constraints}
% \begin{figure}[h]
% \centering
% \includegraphics[scale=0.30]{pictures/acceleration.png}
% \caption{Approximation of Longitudinal Acceleration on vertex $v$.}
% \label{fig:acceleration}
% \end{figure}
To further enhance passenger comfort, longitudinal accelerations need to be considered and penalized. We perform the analysis of accelerations based on discussions over the change of velocities around vertices and establish relevant constraints through linear approximation. In particular, for a vertex $v$ passed through by vehicle $i$, the associated acceleration can be approximately calculated as
\begin{equation}
\label{numericalAcc}
    \frac{d^2s}{dt^2}\bigg|_{t=t^i_v}\approx\frac{l_\beta/(t^i_{\beta_2}-t^i_v)-l_\alpha/(t^i_{v}-t^i_{\alpha_1})}{(t^i_{\beta_2}-t^i_v)/2+(t^i_{v}-t^i_{\alpha_1})/2}\doteq a^i_v,
\end{equation}
where $\alpha$ is the incoming edge of vertex $v$ passed through by vehicle $i$, and $\beta$ is the outgoing edge. (\ref{numericalAcc}) is the central finite difference approximation for the second-order derivative around $t^i_v$. We assume that such approximation holds valid over the time interval $[(t^i_{\alpha_1}+t^i_v)/2,(t^i_v+t^i_{\beta_2})/2]$. Therefore, we propose to minimize $|a^i_v(t^i_{\beta_2}-t^i_{\alpha_1})/2|$. In particular, $a^i_v(t^i_{\beta_2}-t^i_{\alpha_1})/2\doteq\xi^i_{v,\alpha,\beta}$ is given by 
\begin{equation}
\label{accumulatedACC}
\begin{aligned}
    \xi^i_{v,\alpha,\beta} &= l_\beta/(t^i_{\beta_2}-t^i_v)-l_\alpha/(t^i_{v}-t^i_{\alpha_1})\\
    & = V^i_\beta - V^i_\alpha = 1/\Omega^i_\beta- 1/\Omega^i_\alpha,
\end{aligned}
\end{equation}
where $V^i_\alpha$ and $V^i_\beta$ are the average velocities of $i$ on edge $\alpha$ and $\beta$, and $\Omega^i_\alpha = 1/V^i_\alpha, \Omega^i_\beta = 1/V^i_\beta$. It should be noted that although actual accelerations over $[(t^i_{\alpha_1}+t^i_v)/2,(t^i_v+t^i_{\beta_2})/2]$ might fluctuate and deviate away from $a^i_v$, the integration of actual accelerations over time will still accumulate to the change of velocities, and therefore $\xi^i_{v,\alpha,\beta}=V^i_\beta - V^i_\alpha$ can serve as a good estimate of the total amount of acceleration exerted over $[(t^i_{\alpha_1}+t^i_v)/2,(t^i_v+t^i_{\beta_2})/2]$.

Noted that (\ref{accumulatedACC}) is nonlinear with respect to the optimization variables, namely the time stamps. For linearization, we denote the average velocity along the two edges $\alpha$ and $\beta$ as $\Tilde{V}^i_{\alpha\beta}=(l_\alpha+l_\beta)/(t^i_{\beta_2}-t^i_{\alpha_1})$, and the inverse of the average velocity $\Tilde{\Omega}^i_{\alpha\beta}=1/\Tilde{V}^i_{\alpha\beta}$. We then perform linearization of $\xi^i_{v,\alpha,\beta}$ around $\Tilde{\Omega}^i_{\alpha\beta}$, namely
\begin{equation}
\label{linearAcc}
    \xi^i_{v,\alpha,\beta}\approx\frac{\partial\xi^i_{v,\alpha,\beta}}{\partial\Omega^i_\alpha}\bigg|_{\Tilde{\Omega}^i_{\alpha\beta}}(\Omega^i_\alpha-\Tilde{\Omega}^i_{\alpha\beta})+\frac{\partial \xi^i_{v,\alpha,\beta}}{\partial\Omega^i_\beta}\bigg|_{\Tilde{\Omega}^i_{\alpha\beta}}(\Omega^i_\beta-\Tilde{\Omega}^i_{\alpha\beta}).
\end{equation}
In particular,
\begin{equation}
    \frac{\partial\xi^i_{v,\alpha,\beta}}{\partial\Omega^i_\alpha}\bigg|_{\Tilde{\Omega}^i_{\alpha\beta}}=(\Tilde{V}^{i}_{\alpha\beta})^2,
    \frac{\partial\xi^i_{v,\alpha,\beta}}{\partial\Omega^i_\beta}\bigg|_{\Tilde{\Omega}^i_{\alpha\beta}}=-(\Tilde{V}^{i}_{\alpha\beta})^2.
\end{equation}
Plug it back into (\ref{linearAcc}) and we have
\begin{equation}
\begin{aligned}
\label{linearACC2}
\xi^i_{v,\alpha,\beta}&\approx(\Tilde{V}^{i}_{\alpha\beta})^2(\Omega^i_\alpha-\Omega^i_\beta)=(\Tilde{V}^{i}_{\alpha\beta})^2(\frac{t^i_v-t^i_{\alpha_1}}{l_\alpha}-\frac{t^i_{\beta_2}-t^i_v}{l_\beta}).
\end{aligned}
\end{equation}
In (\ref{linearACC2}), the nonlinearity with respect to time stamps is incurred by $\Tilde{V}^i_{\alpha\beta}$, which is an optimization variable that depends on both $t^i_{\beta_2}$ and $t^i_{\alpha_1}$. An intuitive solution to render (\ref{linearACC2}) linear is to replace $\Tilde{V}^i_{\alpha\beta}$ with $V^i_r$. This approximation works well when $\Tilde{V}^i_{\alpha\beta}\approx V^i_r$, but becomes poor when the velocity of CAV $i$ deviates from $V^i_r$. Due to the rapid surge of the quadratic function, errors up to orders of magnitude can be induced.

To cope with this problem, we first divide the allowed velocity region $[V^i_{slow},V^i_{fast}]$ into a set of mutually exclusive regions $[V^{i,k}_{slow}, V^{i,k}_{fast}]_{k\in\mathcal{K}}$. For each region $[V^{i,k}_{slow}, V^{i,k}_{fast}]$, we adopt a reference velocity $V^{i,k}_r\in[V^{i,k}_{slow}, V^{i,k}_{fast}]$ such that when $\Tilde{V}^i_{\alpha\beta}\in[V^{i,k}_{slow}, V^{i,k}_{fast}]$, this pre-assigned velocity $V^{i,k}_r$ will be used as the linearization operating point instead. The problem then becomes to identify the velocity region that $\Tilde{V}^i_{\alpha\beta}$ belongs to. For this purpose, we define a set of binary variables $\{m^{i,k}_{v,\alpha,\beta}\}_{k\in\mathcal{K}}$ such that when $\Tilde{V}^i_{\alpha\beta}\in [V^{i,\Tilde{k}}_{slow}, V^{i,\Tilde{k}}_{fast}]$, the corresponding $m^{i,\Tilde{k}}_{v,\alpha,\beta}$ is equal to 1 and all others are equal to zero. The following constraints are introduced to enforce this condition:
\begin{subequations}
\label{identifyRegion}
\begin{align}
&\sum_{k\in\mathcal{K}}m^{i,k}_{v,\alpha,\beta}=1,\label{identifyRegionA}\\
&\sum_{k\in\mathcal{K}}m^{i,k}_{v,\alpha,\beta}\frac{l_\alpha+l_\beta}{V^{i,k}_{slow}}\geq t^i_{\beta_2}-t^i_{\alpha_1}-M(2-y^i_\alpha-y^i_\beta),\label{identifyRegionB}\\
&\sum_{k\in\mathcal{K}}m^{i,k}_{v,\alpha,\beta}\frac{l_\alpha+l_\beta}{V^{i,k}_{fast}}\leq t^i_{\beta_2}-t^i_{\alpha_1}+M(2-y^i_\alpha-y^i_\beta),\label{identifyRegionC}\\
&\forall i\in\mathcal{N},v\in\Bar{\mathcal{V}}^i,\alpha\in\mathcal{E}^{i,in}_v,\beta\in\mathcal{E}^{i,out}_v.\nonumber
\end{align}
\end{subequations}
The principle behind the above constraints is that for each set of $\{m^{i,k}_{v,\alpha,\beta}\}_{k\in\mathcal{K}}$, only one of them is equal to $1$ and all others are equal to $0$, as is guaranteed by (\ref{identifyRegionA}). When $\Tilde{V}^i_{\alpha\beta}\in [V^{i,\Tilde{k}}_{slow}, V^{i,\Tilde{k}}_{fast}]$, the enforcement of the following constraint
\begin{equation}
\label{identifyRegionRaw}
    \sum_{k\in\mathcal{K}}m^{i,k}_{v,\alpha,\beta}V^{i,k}_{slow}\leq \Tilde{V}^i_{\alpha\beta} \leq \sum_{k\in\mathcal{K}}m^{i,k}_{v,\alpha,\beta}V^{i,k}_{fast}
\end{equation}
is sufficient to guarantee that
\begin{equation}
    \sum_{k\in\mathcal{K}}m^{i,k}_{v,\alpha,\beta}V^{i,k}_{slow}= V^{i,\Tilde{k}}_{slow}, \sum_{k\in\mathcal{K}}m^{i,k}_{v,\alpha,\beta}V^{i,k}_{fast}= V^{i,\Tilde{k}}_{fast},
\end{equation}
which then ensures $m^{i,\Tilde{k}}_{v,\alpha,\beta}=1$ and $m^{i,k}_{v,\alpha,\beta}=0$ for $k\neq\Tilde{k}$. Thus, the region that $\Tilde{V}^i_{\alpha\beta}$ belongs to is identified. Simple transformation of (\ref{identifyRegionRaw}) results in (\ref{identifyRegionB}) and (\ref{identifyRegionC}) which are linear with respect to time stamps. Noted that the big-M method is again utilized to ensure that both (\ref{identifyRegionB}) and (\ref{identifyRegionC}) are activated only if $y^i_\alpha=y^i_\beta=1$, namely $\Tilde{V}^i_{\alpha\beta}$ is valid.

Similar to velocity constraints, we again apply the big-M method to indicate whether an acceleration constraint is activated, and we also introduce a pair of positive slack variables $\{\Bar{\gamma}^{i,k}_{v,\alpha,\beta},\underline{\gamma}^{i,k}_{v,\alpha,\beta}\}$ to penalize acceleration and deceleration. The resulting constraints are given as follows:
\begin{align}\label{acceleration}
    &\frac{t^i_v-t^i_{\alpha_1}}{l_\alpha}-\frac{t^i_{\beta_2}-t^i_v}{l_\beta}\leq M(3-y^i_\alpha-y^i_\beta-m^{i,k}_{v,\alpha,\beta})+\Bar{\gamma}^{i,k}_{v,\alpha,\beta},\nonumber\\
    &\frac{t^i_v-t^i_{\alpha_1}}{l_\alpha}-\frac{t^i_{\beta_2}-t^i_v}{l_\beta}\geq -M(3-y^i_\alpha-y^i_\beta-m^{i,k}_{v,\alpha,\beta})-\underline{\gamma}^{i,k}_{v,\alpha,\beta},\nonumber\\
    &0\leq\Bar{\gamma}^{i,k}_{v,\alpha,\beta}\leq\frac{\gamma_{max}(t^i_{\beta_2}-t^i_{\alpha_1})}{2(V^{i,k}_r)^2}+M(3-y^i_\alpha-y^i_\beta-m^{i,k}_{v,\alpha,\beta}),\nonumber\\
    &0\leq\underline{\gamma}^{i,k}_{v,\alpha,\beta}\leq\frac{-\gamma_{min}(t^i_{\beta_2}-t^i_{\alpha_1})}{2(V^{i,k}_r)^2}+M(3-y^i_\alpha-y^i_\beta-m^{i,k}_{v,\alpha,\beta}),\nonumber\\
    &\forall k\in\mathcal{K},i\in\mathcal{N},v\in\Bar{\mathcal{V}}^i,\alpha\in\mathcal{E}^{i,in}_v,\beta\in\mathcal{E}^{i,out}_v.
\end{align}
Similarly, the above constraints will be activated only if $y^i_\alpha=y^i_\beta=m^{i,k}_{v,\alpha,\beta}=1$, namely when $\Tilde{V}^i_{\alpha\beta}$ is valid and $\Tilde{V}^i_{\alpha\beta}\in [V^{i,k}_{slow}, V^{i,k}_{fast}]$. $\gamma_{max}$ and $\gamma_{min}$ are a pair of pre-defined constants to prevent the accelerations from going unbounded. Through adding $\Bar{\gamma}^{i,k}_{v,\alpha,\beta}$ and $\underline{\gamma}^{i,k}_{v,\alpha,\beta}$ to the overall cost, the integration of longitudinal accelerations can be penalized.

The acceleration constraints on the starting vertices are essentially different from those for intermediate vertices, given the fact that the initial velocities are already known and fixed. We define the initial velocity for CAV $i$ as $V^i_{init}$. The constraints for identifying the velocity region of $V^i_\beta$ with $\beta\in\mathcal{E}^{i,out}_{s^i}$ are given as
\begin{equation}
\label{initIdentifyRegion}
\begin{aligned}
    &\sum_{k\in\mathcal{K}}m^{i,k}_{s^i,\beta}=1,\\
    &\sum_{k\in\mathcal{K}}m^{i,k}_{s^i,\beta}\frac{l_\beta}{V^{i,k}_{slow}}\geq t^i_{\beta_2}-t^i_{s^i}-M(1-y^i_\beta),\\
&\sum_{k\in\mathcal{K}}m^{i,k}_{s^i,\beta}\frac{l_\beta}{V^{i,k}_{fast}}\leq t^i_{\beta_2}-t^i_{s^i}+M(1-y^i_\beta),\\
&\forall i\in\mathcal{N},\beta\in\mathcal{E}^{i,out}_{s^i}.
\end{aligned}
\end{equation}
Given the finite difference approximation 
\begin{equation}
\label{numericalAccInit}
    \frac{d^2s}{dt^2}\bigg|_{t=t^i_{s^i}}\approx\frac{l_\beta/(t^i_{\beta_2}-t^i_{s^i})-V^i_{init}}{(t^i_{\beta_2}-t^i_{s^i})/2}\doteq a^i_{s^i},
\end{equation}
the constraints imposed on $a^i_{s^i}(t^i_{\beta_2}-t^i_{s^i})/2\doteq\xi^i_{s^i,\beta}$ is derived as
\begin{align}\label{initAcceleration}
&\frac{2V^{i,k}_r-V^i_{init}}{(V^{i,k}_r)^2}-\frac{t^i_{\beta_2}-t^i_{s^i}}{l_\beta}\leq M(2-y^i_\beta-m^{i,k}_{s^i,\beta})+\Bar{\gamma}^{i,k}_{s^i,\beta},\nonumber\\
&\frac{2V^{i,k}_r-V^i_{init}}{(V^{i,k}_r)^2}-\frac{t^i_{\beta_2}-t^i_{s^i}}{l_\beta}\geq -M(2-y^i_\beta-m^{i,k}_{s^i,\beta})-\underline{\gamma}^{i,k}_{s^i,\beta},\nonumber\\
&0\leq\Bar{\gamma}^{i,k}_{s^i,\beta}\leq\frac{\gamma_{max}(t^i_{\beta_2}-t^i_{s^i})}{2(V^{i,k}_r)^2}+M(2-y^i_\beta-m^{i,k}_{s^i,\beta}),\nonumber\\
&0\leq\underline{\gamma}^{i,k}_{s^i,\beta}\leq\frac{-\gamma_{min}(t^i_{\beta_2}-t^i_{s^i})}{2(V^{i,k}_r)^2}+M(2-y^i_\beta-m^{i,k}_{s^i,\beta}),\nonumber\\
&\forall k\in\mathcal{K}, i\in\mathcal{N},\beta\in\mathcal{E}^{i,out}_{s^i}.
\end{align}
Noted that $\Bar{\gamma}^{i,k}_{s^i,\beta}$ and $\underline{\gamma}^{i,k}_{s^i,\beta}$ are also a pair of slack variables that are similarly included in the overall cost to penalize the boundary accelerations. The derivation is similar and therefore omitted.

\subsubsection{Steering Effects Constraints}
To avoid obtaining a zigzag path or a path that requires vehicles to take sharp turns at high velocities, the steering effect must be taken into consideration. In most urban traffic scenarios, the main factor that prevents vehicles from taking sharp turns is the lateral acceleration induced by steering, which not only impairs the comfort of passengers but can also lead to potential danger. Therefore, we propose to penalize the integration of lateral acceleration over the entire time span. In the proposed method, steering will only take place around vertices. For a vertex $v$ and $\alpha, \beta$ as its incoming and outgoing edges respectively, we define the angle between $\alpha$ and $\beta$ as $\theta_{\alpha,\beta}$. For CAV $i$ that passes through both $\alpha$ and $\beta$, the total amount of lateral acceleration induced by steering at vertex $v$ is estimated as
\begin{equation}
\begin{aligned}
    \int_{t^i_{\alpha_1}}^{t^i_{\beta_2}}a^i_{lat}(t)dt &= \int_{t^i_{\alpha_1}}^{t^i_{\beta_2}}V^i(t)\omega ^i(t)dt\\
    &\approx \Tilde{V}^{i}_{\alpha\beta}\int_{t^i_{\alpha_1}}^{t^i_{\beta_2}}\omega ^i(t)dt=\Tilde{V}^{i}_{\alpha\beta}\theta_{\alpha,\beta},
\end{aligned}
\end{equation}
where $a^i_{lat}$ is the lateral acceleration, $V^i$ is the velocity, and $\omega ^i$ is the angular velocity. In the above equation, we approximate the velocity over the time interval $[t^i_{\alpha_1},t^i_{\beta_2}]$ with the average velocity $\Tilde{V}^{i}_{\alpha\beta}$. It is obvious that the summation of $\Tilde{V}^{i}_{\alpha\beta}\theta_{\alpha,\beta}$ over all the vertices passed through by CAV $i$ constitute a good estimator of the total amount of lateral acceleration underwent by CAV $i$ during the entire trip. Noted that $\Tilde{V}^{i}_{\alpha\beta}$ is nonlinear with respect to time stamps, we again replace $\Tilde{V}^{i}_{\alpha\beta}$ with the corresponding $V^{i,k}_r$. The following constraints are then imposed:
\begin{align}\label{steering}
    &V^{i,k}_r\theta_{\alpha,\beta}-M(3-y^i_\alpha-y^i_\beta-m^{i,k}_{v,\alpha,\beta})\leq \eta^{i,k}_{v,\alpha,\beta},\nonumber\\
    &0\leq\eta^{i,k}_{v,\alpha,\beta}\leq \eta_{max}(t^i_{\beta_2}-t^i_{\alpha_1})+M(3-y^i_\alpha-y^i_\beta-m^{i,k}_{v,\alpha,\beta}),\nonumber\\
    &\forall k\in\mathcal{K},i\in\mathcal{N},v\in\Bar{\mathcal{V}}^i,\alpha\in\mathcal{E}^{i,in}_v,\beta\in\mathcal{E}^{i,out}_v.
\end{align}
$\eta^{i,k}_{v,\alpha,\beta}$ is the slack variable to be penalized, and $\eta_{max}$ is a pre-defined and fixed constant to bound the lateral accelerations and to disable sharp turns for CAVs running at relatively high velocities. Corresponding constraints on the starting vertex are given as
\begin{align}\label{initSteering}
    &V^{i,k}_r\theta^i_{init,\beta}-M(3-y^i_\alpha-y^i_\beta-m^{i,k}_{s^i,\beta})\leq \eta^{i,k}_{s^i,\beta},\nonumber\\
    &0\leq\eta^{i,k}_{s^i,\beta}\leq \eta_{max}(t^i_{\beta_2}-t^i_{s^i})+M(3-y^i_\alpha-y^i_\beta-m^{i,k}_{s^i,\beta}),\nonumber\\
    &\forall k\in\mathcal{K}, i\in\mathcal{N},\beta\in\mathcal{E}^{i,out}_{s^i}.
\end{align}
In particular, $\theta^i_{init,\beta}$ is the angle between initial heading of CAV $i$ and edge $\beta$.

\subsubsection{Collision Avoidance Constraints}
To ensure driving safety, constraints must be imposed to avoid possible collisions between any pairs of CAVs. For collision avoidance, we first introduce the concept of critical edge pairs. For a vehicle $i\in\mathcal{N}$ and $e^i\in\mathcal{E}^i$, we consider the case when $i$ is aligned with $e^i$, and $i$ starts from the head of $e^i$ and moves along the edge till the tail of $e^i$. We define the volume swept by this process as $\mathcal{S}^i_{e^i}$. With this notation, we introduce the definition of critical edge pairs.

\begin{definition}
For $i,j\in\mathcal{N}, i\neq j$, $e^i\in\mathcal{E}^i$, and $e^j\in\mathcal{E}^j$, $\{(i,e^i),(j,e^j)\}$ forms a critical edge pairs if $S^i_{e^i}\cap S^j_{e^j}\neq 0$. The set of all such critical edge pairs is denoted as $\Gamma$.
\end{definition}

\begin{figure}[t]
\centering
\includegraphics[scale=0.21]{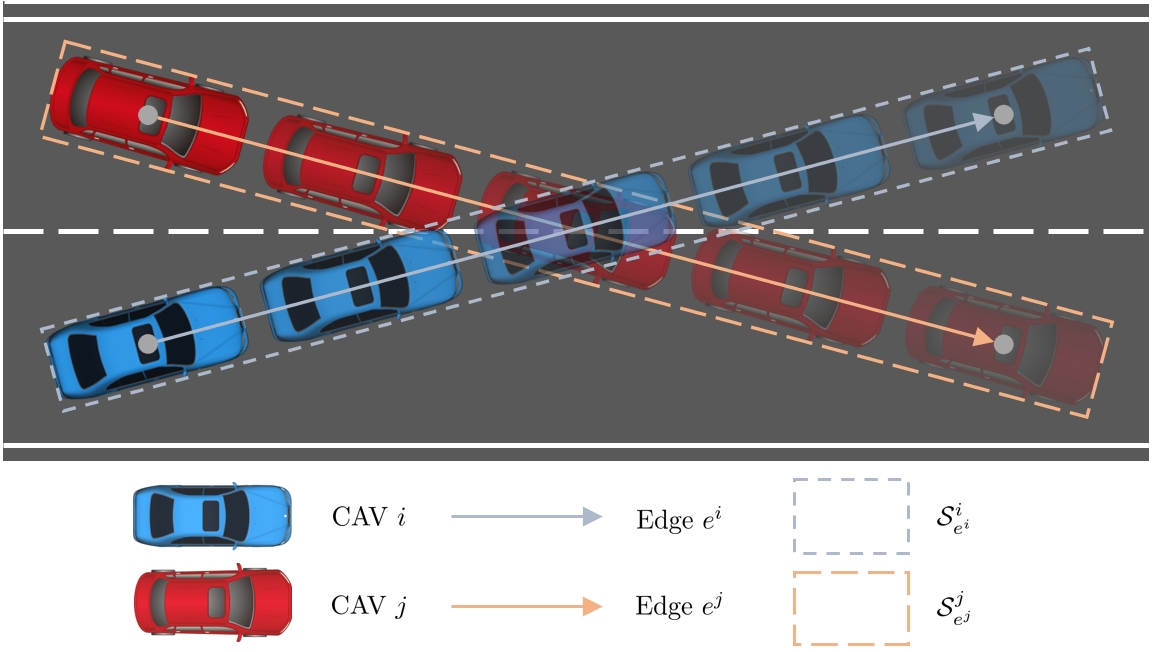}
\caption{An illustrative example of critical edge pairs. Collision may occur when CAV $i$ moves along edge $e^i$ and CAV $j$ moves along edge $e^j$ at the same time.}
\label{fig:criticalPairs}
\end{figure}

An illustrative example of critical edge pairs is given in Fig. \ref{fig:criticalPairs}. Note that the definition of critical edge pairs naturally includes the cases when $e^i=e^j$. Consequently, a sufficient condition for collision avoidance is that the time spans for CAVs to pass through critical edge pairs should be completely non-overlapping. Formally, for any $\{(i,e^i),(j,e^j)\}$ that constitutes a critical edge-pair, we have:
\begin{itemize}
    \item $t^i_{e^i_2}-t^j_{e^j_1}\leq 0$, OR 
    \item $t^j_{e^j_2}-t^i_{e^i_1}\leq 0$, OR
    \item $y^i_{e^i}+y^j_{e^j}\leq 1$.
\end{itemize}

However, it is clear that such collision avoidance constraints can be overly conservative, especially in cases when the lengths of edges are relatively long. The overall performance of the proposed method will then rely heavily on the construction of the waypoint graph and the lengths of the edges, which is certainly undesirable. To avoid the conservativeness, we further identify the critical region on each of the edges in critical edge pairs where possible collisions may occur. Formally, we define $p_v$ as the location of vertex $v$, and $\mathcal{S}^i_{e^i}(\theta)$ as the volume occupied by vehicle $i$ when it is aligned with edge $i$ and its location is at $(1-\theta)p_{e^i_1}+\theta p_{e^i_2}$. Then we identify the critical region $\mathcal{C}^{ij}_{e^ie^j}$ on edge $i$ as
\begin{equation}
\begin{aligned}
    \theta^{ij,1}_{e^ie^j}&= \min_{0\leq\theta\leq1}\{\mathcal{S}^i_{e^i}(\theta)\cap\mathcal{S}^j_{e^j}\neq 0\},\\
    \theta^{ij,2}_{e^ie^j}&= \max_{0\leq\theta\leq1}\{\mathcal{S}^i_{e^i}(\theta)\cap\mathcal{S}^j_{e^j}\neq 0\},\\
    \mathcal{C}^{ij}_{e^ie^j}&= \{(1-\theta)p_{e^i_1}+\theta p_{e^i_2}|\theta^{ij,1}_{e^ie^j}\leq\theta\leq\theta^{ij,2}_{e^ie^j}\}.
\end{aligned}
\end{equation}

\begin{figure}[t]
\centering
\includegraphics[scale=0.1475]{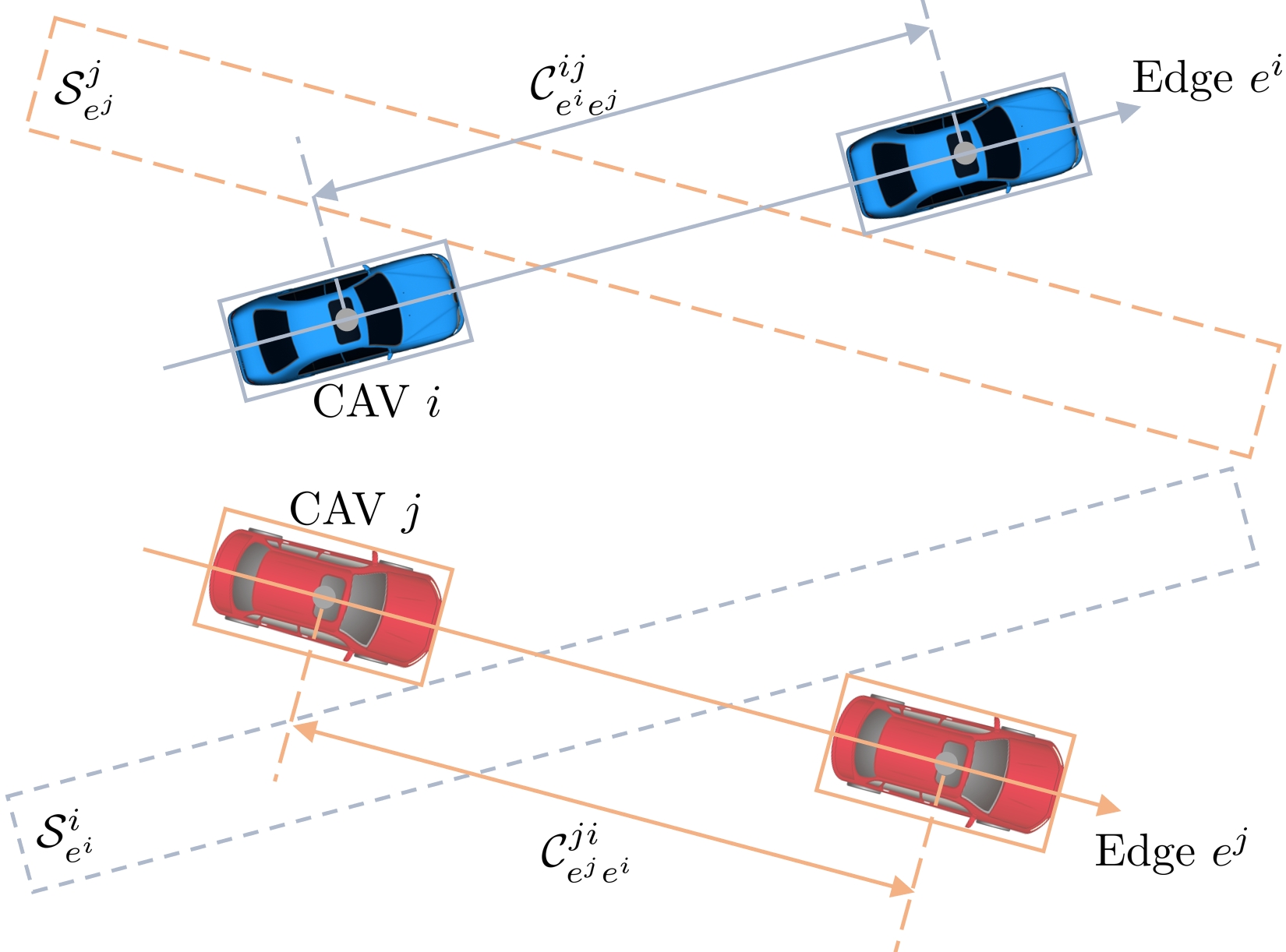}
\caption{An illustrative example of critical regions within a critical edge pair. Collision may occur when both CAVs enter the corresponding critical region at the same time.}
\label{fig:criticalRegion}
\end{figure}

An illustrative example of critical regions is shown in Fig. \ref{fig:criticalRegion}. It is obvious that $\mathcal{C}^{ij}_{e^ie^j}$ can be determined before the optimization process. We also denote the time stamp when vehicle $i$ enters $\mathcal{C}^{ij}_{e^ie^j}$ as $t^{ij,1}_{e^ie^j}$ and the time stamp when it leaves $\mathcal{C}^{ij}_{e^ie^j}$ as $t^{ij,2}_{e^ie^j}$. Similar definitions are made for vehicle $j$ entering the critical region $\mathcal{C}^{ji}_{e^je^i}$
on edge $e^j$. With the assumption that the motions of vehicles along edges are approximately uniform, we obtain the following approximations through linear interpolation: 
\begin{equation}
\label{linearInterpolation}
\begin{aligned}
    &t^{ij,1}_{e^ie^j} \approx (1-\theta^{ij,1}_{e^ie^j})t^i_{e^i_1}+\theta^{ij,1}_{e^ie^j} t^i_{e^i_2},\\
    &t^{ij,2}_{e^ie^j} \approx (1-\theta^{ij,2}_{e^ie^j})t^i_{e^i_1}+\theta^{ij,2}_{e^ie^j} t^i_{e^i_2},\\
    &t^{ji,1}_{e^je^i} \approx (1-\theta^{ji,1}_{e^je^i})t^j_{e^j_1}+\theta^{ji,1}_{e^je^i} t^j_{e^j_2},\\
    &t^{ji,2}_{e^je^i} \approx (1-\theta^{ji,2}_{e^je^i})t^j_{e^j_1}+\theta^{ji,2}_{e^je^i} t^j_{e^j_2}.\\   
\end{aligned}
\end{equation}

\begin{figure}[t]
\centering
\includegraphics[scale=0.162]{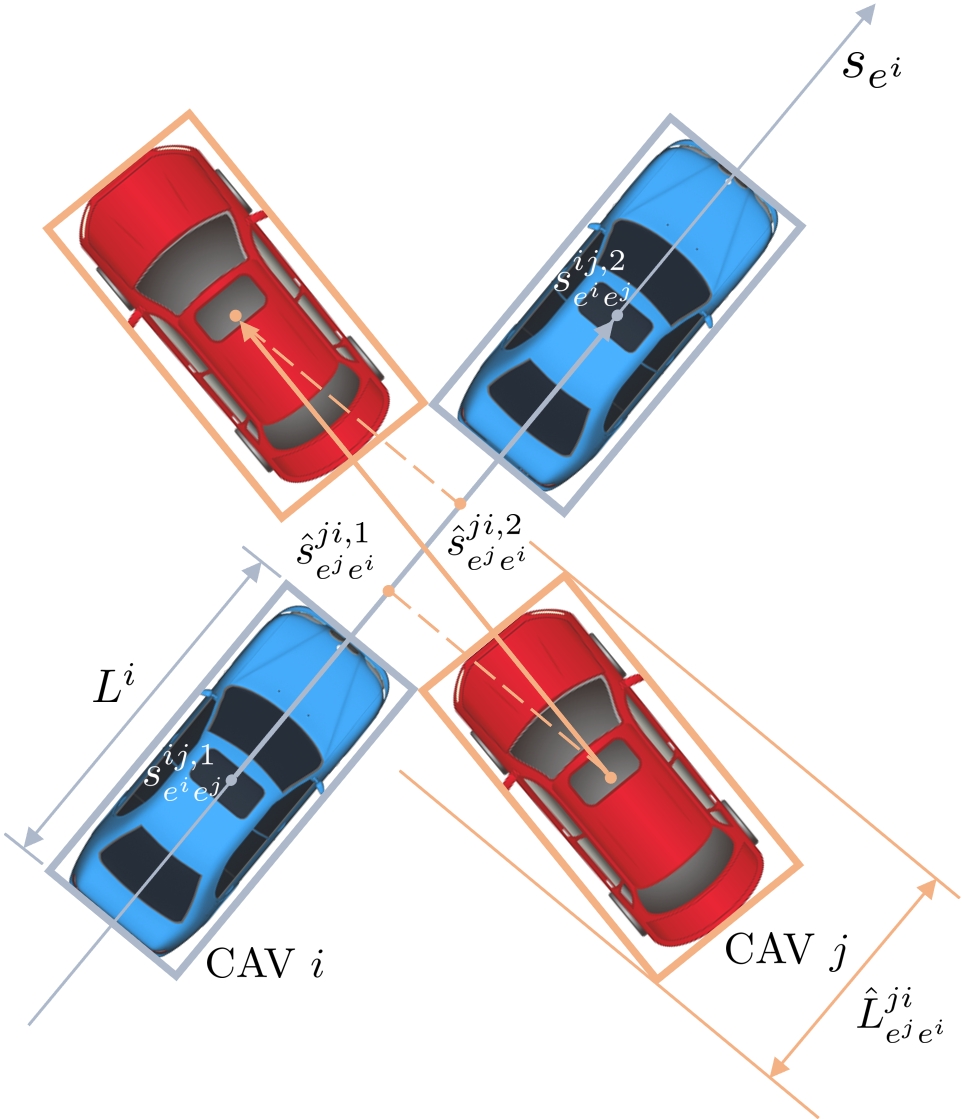}
\caption{Projection of motion of CAV $j$ onto edge $e^i$.}
\label{fig:projection}
\end{figure}

It should be noted that the critical edge pairs and critical regions include many different situations. To derive necessary conditions for collision avoidance under all these situations, we first project the two-dimensional motions of vehicle $i$ and vehicle $j$ along edge $e^i$ and $e^j$ into one dimension, and conclude that the distance between the projection of these two CAVs should be greater than an equivalent safety distance. Without loss of generality, for a critical edge pair $\{(i,e^i),(j,e^j)\}$, we project the motion of vehicle $j$ along edge $e^j$ onto edge $e^i$, as shown in  Fig. \ref{fig:projection}. We first construct a one-dimensional coordinate system $s_{e^i}$ that is aligned with edge $i$. We then project both $\mathcal{C}^{ij}_{e^ie^j}$ and $\mathcal{C}^{ji}_{e^je^i}$ onto $s_{e^i}$. As is shown in Fig. \ref{fig:projection}, the coordinates of starting point and end point of $\mathcal{C}^{ij}_{e^ie^j}$ are denoted as $s^{ij,1}_{e^ie^j}$ and $s^{ij,2}_{e^ie^j}$, respectively, and the coordinates of the projections of starting point and end point of $\mathcal{C}^{ji}_{e^je^i}$ are denoted as $\hat{s}^{ji,1}_{e^je^i}$ and $\hat{s}^{ji,2}_{e^je^i}$. $L^i$ is the length of CAV $i$ and $\hat{L}^{ji}_{e^je^i}$ is the length of projection of CAV $j$ onto coordinate system $s_{e^i}$. Moreover, the equivalent safety distance is given as $D^{ij}_{e^ie^j}=\frac{1}{2}(L^i+\hat{L}^{ji}_{e^je^i})$. With the above transformations and definitions, the collision avoidance condition then turns into keeping a safety distance $D^{ij}_{e^ie^j}$ between the center of CAV $i$ and the projection of the center of CAV $j$, while the center of CAV $i$ is moving from $s^{ij,1}_{e^ie^j}$ to $s^{ij,2}_{e^ie^j}$ and the projection of the center of CAV $j$ is moving from $\hat{s}^{ji,1}_{e^je^i}$ to $\hat{s}^{ji,2}_{e^je^i}$. To derive the linear collision avoidance constraints, we consider two different cases:
\begin{itemize}
    \item \textit{Case 1: the angle between $e^i$ and $e^j$ is smaller than $\frac{\pi}{2}$.}
    \item \textit{Case 2: the angle between $e^i$ and $e^j$ is greater than or equal to $\frac{\pi}{2}$.}
\end{itemize}
For each of the cases, we discuss the situations whether CAV $i$ is passing through the critical region after or before CAV $j$, namely:
\begin{itemize}
    \item \textit{Case 1.1: the angle between $e^i$ and $e^j$ is smaller than $\frac{\pi}{2}$, and CAV $i$ is passing through the critical region after CAV $j$.}
    \item \textit{Case 1.2: the angle between $e^i$ and $e^j$ is smaller than $\frac{\pi}{2}$, and CAV $i$ is passing through the critical region before CAV $j$.}
    \item \textit{Case 2.1: the angle between $e^i$ and $e^j$ is greater than or equal to $\frac{\pi}{2}$, and CAV $i$ is passing through the critical region after CAV $j$.}
    \item \textit{Case 2.2: the angle between $e^i$ and $e^j$ is greater than or equal to $\frac{\pi}{2}$, and CAV $i$ is passing through the critical region before CAV $j$.}
\end{itemize}

Before we go into detailed discussions of the above cases, we need to introduce a pair of binary variables $\delta^{ij}_{e^ie^j}$ and $\delta^{ji}_{e^je^i}$ to determine the passing order of CAV $i$ and $j$. These two binary variables are designed such that when $\delta^{ij}_{e^ie^j}=\delta^{ji}_{e^je^i}=0$, at most one of the two CAVs is actually passing through the critical edge pairs. When $\delta^{ij}_{e^ie^j}=0$ and $\delta^{ji}_{e^je^i}=1$, CAV $i$ is supposed to pass through after CAV $j$, corresponding to \textit{Case 1.1} and \textit{Case 2.1}. When $\delta^{ij}_{e^ie^j}=1$ and $\delta^{ji}_{e^je^i}=0$, CAV $i$ is supposed to pass through before CAV $j$, corresponding to \textit{Case 1.2} and \textit{Case 2.2}. It is illegal to have $\delta^{ij}_{e^ie^j}=\delta^{ji}_{e^je^i}=1$. As a result, we have the following constraints imposed on all cases:
\begin{equation}
\label{Collision1}
\begin{aligned}
y^i_{e^i}+y^j_{e^j}-M(\delta^{ij}_{e^ie^j}+\delta^{ji}_{e^je^i})&\leq 1,\\
\delta^{ji}_{e^je^i}+\delta^{ij}_{e^ie^j}\ \ \ \ \ \ \ \ \ &\leq 1.
\end{aligned}
\end{equation}

% \begin{figure}[t]

% \centering
% \subfigure[]{\includegraphics[scale=0.425]{pictures/Case 1.1 (b).png}}
% \subfigure[]{\includegraphics[scale=0.425]{pictures/Case 1.1 (c).png}}

% \caption{$t$-$s$ curves of Case 1.1.}
% \label{fig:Case1.1}
% \end{figure}

% \begin{figure}[t]

% \centering
% \subfigure[]{\includegraphics[scale=0.425]{pictures/Case 1.2 (a).png}}
% \subfigure[]{\includegraphics[scale=0.425]{pictures/Case 1.2 (b).png}}

% \caption{$t$-$s$ curves of Case 1.2.}
% \label{fig:Case1.2}
% \end{figure}

\begin{figure*}[t]
\centering
\subfigure[Condition 1 of Case 1.1]{\includegraphics[scale=0.345]{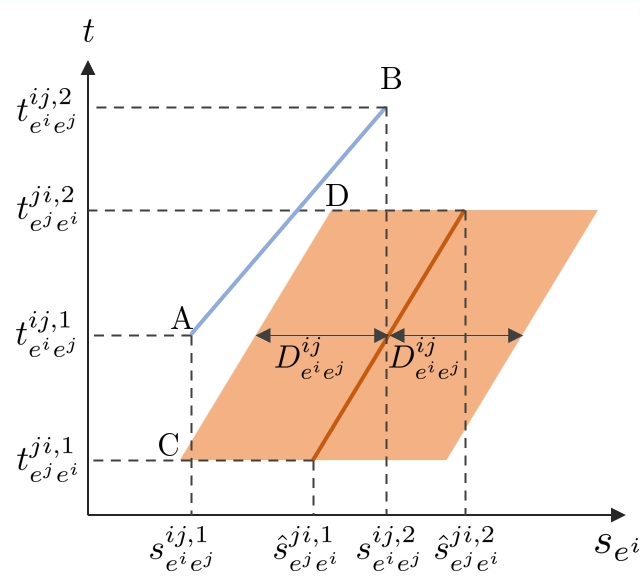}}
\subfigure[Condition 2 of Case 1.1]{\includegraphics[scale=0.345]{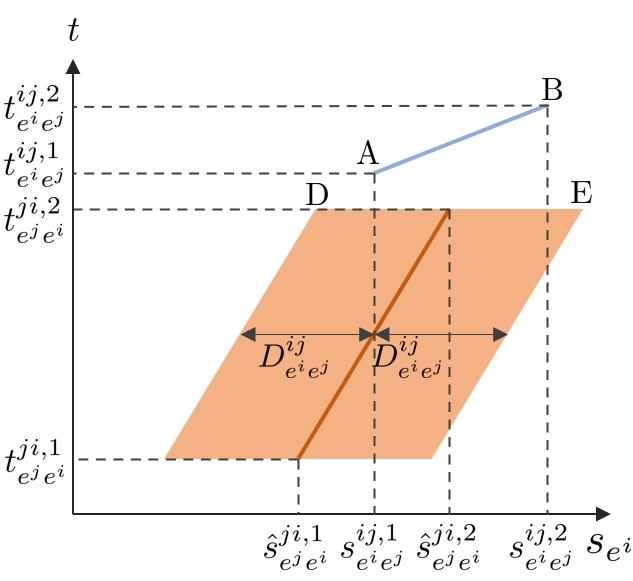}}
\subfigure[Condition 1 of Case 1.2]{\includegraphics[scale=0.345]{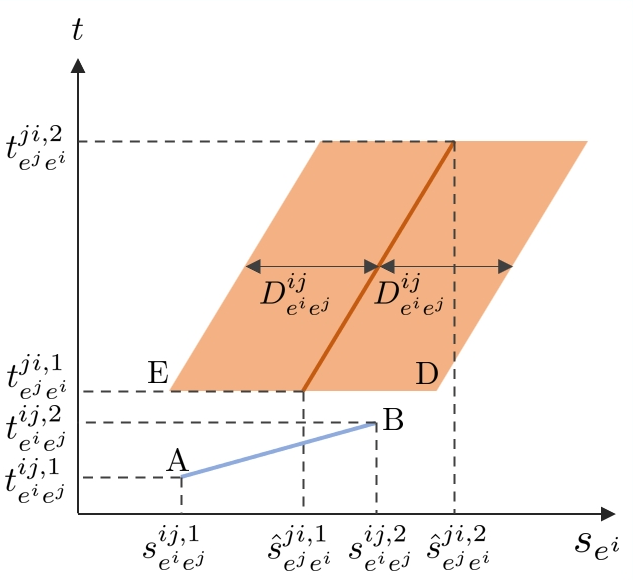}}
\subfigure[Condition 2 of Case 1.2]{\includegraphics[scale=0.345]{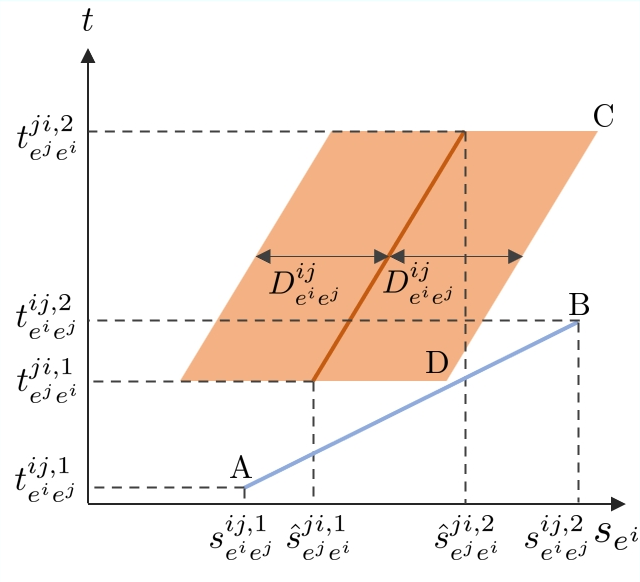}}
\subfigure[Case 2.1]{\includegraphics[scale=0.345]{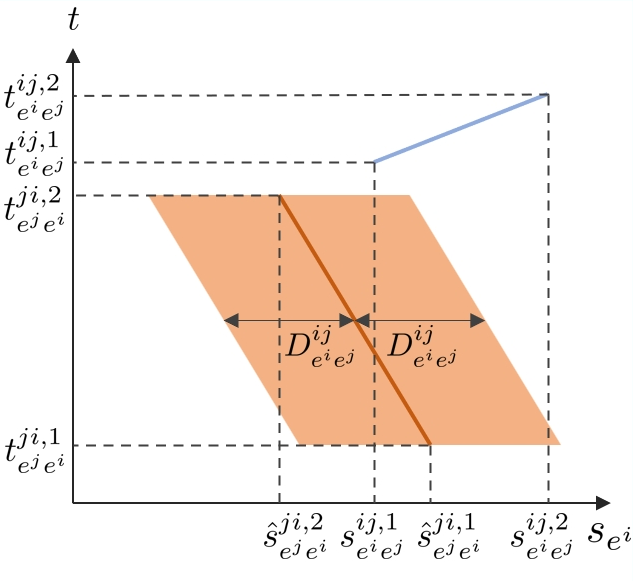}}
\subfigure[Case 2.2]{\includegraphics[scale=0.345]{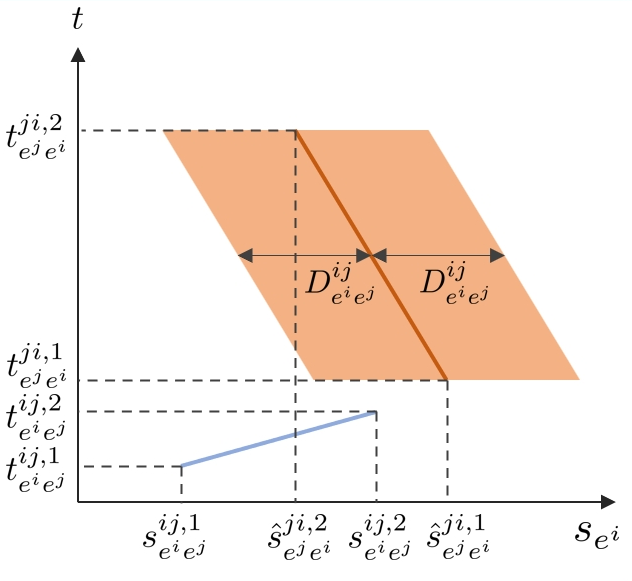}}

\caption{t-s curves of different cases and conditions.}

\label{fig:collision}
\end{figure*}

\textit{Case 1.1:} To derive the linear constraints for keeping safety distance between the two CAVs in this case, we examine the $t$-$s$ curves of both CAVs, which are illustrated in Fig. \ref{fig:collision}(a)(b). In those figures, the orange line segments demonstrate the $t$-$s$ curves of CAV $j$ while passing through the critical region of edge $e^j$, and the blue line segments are the corresponding $t$-$s$ curves of CAV $i$. Then the sufficient condition for keeping safety distance between CAV $i$ and $j$ is that the blue line segments should not enter the critical zones spanned by the orange line segments (demonstrated by the orange parallelograms). 

One noticeable fact is that both $s^{ij,1}_{e^ie^j}$ and $s^{ij,2}_{e^ie^j}$ are confined to the region $[\hat{s}^{ji,1}_{e^je^i}-D^{ij}_{e^ie^j}, \hat{s}^{ji,2}_{e^je^i}+D^{ij}_{e^ie^j}]$, or otherwise $s^{ij,1}_{e^ie^j}$ and $s^{ij,2}_{e^ie^j}$ will not belong to the critical region of edge $e^i$. Also, $\hat{s}^{ji,1}_{e^je^i}$ and $\hat{s}^{ji,2}_{e^je^i}$ are confined to the region $[s^{ij,1}_{e^ie^j}-D^{ij}_{e^ie^j}, s^{ji,2}_{e^ie^j}+D^{ij}_{e^ie^j}]$ for the same reason. These facts lead to the following results:
\begin{itemize}
    \item $s^{ij,1}_{e^ie^j}\in[\hat{s}^{ji,1}_{e^je^i}-D^{ij}_{e^ie^j}, \hat{s}^{ji,1}_{e^je^i}+D^{ij}_{e^ie^j}]$, and
    \item $s^{ij,2}_{e^ie^j}\in[\hat{s}^{ji,2}_{e^je^i}-D^{ij}_{e^ie^j}, \hat{s}^{ji,2}_{e^je^i}+D^{ij}_{e^ie^j}]$.
\end{itemize}

Under these results, we discuss the following two conditions:

\textit{Condition 1 of Case 1.1:} When $s^{ij,1}_{e^ie^j}<\hat{s}^{ji,2}_{e^je^i}-D^{ij}_{e^ie^j}$, the sufficient condition for keeping safety distance is that point A should stay above line CD and point D should stay below line AB (see Fig. \ref{fig:collision}(a)). Since Case 1.1 is enforced by setting $\delta^{ji}_{e^je^i}=1$, we have the following constraints:
\begin{equation}
\label{AaboveCD}
\begin{aligned}
    t^{ij,1}_{e^ie^j}\geq&\frac{s^{ij,1}_{e^ie^j}-(\hat{s}^{ji,1}_{e^je^i}-D^{ij}_{e^ie^j})}{\hat{s}^{ji,2}_{e^je^i}-\hat{s}^{ji,1}_{e^je^i}}(t^{ji,2}_{e^je^i}-t^{ji,1}_{e^je^i}) + t^{ji,1}_{e^je^i}\\
    &-M(1-\delta^{ji}_{e^je^i}),\\
\end{aligned}
\end{equation}
and
\begin{equation}
\begin{aligned}
    t^{ji,2}_{e^je^i}\leq&\frac{(\hat{s}^{ji,2}_{e^je^i}-D^{ij}_{e^ie^j})-s^{ij,1}_{e^ie^j}}{s^{ij,2}_{e^ie^j}-s^{ij,1}_{e^ie^j}}(t^{ij,2}_{e^ie^j}-t^{ij,1}_{e^ie^j})+t^{ij,1}_{e^ie^j}\\
    &+M(1-\delta^{ji}_{e^je^i}).
\end{aligned}
\end{equation}

\textit{Condition 2 of Case 1.1:} When $s^{ij,1}_{e^ie^j}\geq\hat{s}^{ji,2}_{e^je^i}-D^{ij}_{e^ie^j}$, the sufficient condition can be easily seen from Fig. \ref{fig:collision}(b), which is
\begin{equation}
\label{AaboveDE}
    t^{ij,1}_{e^ie^j}\geq t^{ji,2}_{e^je^i}-M(1-\delta^{ji}_{e^je^i}).
\end{equation}

\textit{Case 1.2:} The $t$-$s$ curves corresponding to this case are shown in Fig. \ref{fig:collision}(c)(d). Meanwhile, the same conclusions about lower and upper bounds of $s^{ij,1}_{e^ie^j}$ and $s^{ij,2}_{e^ie^j}$ still apply. We also discuss the following two conditions:

\textit{Condition 1 of Case 1.2:} When $s^{ij,2}_{e^ie^j}\leq\hat{s}^{ji,1}_{e^je^i}+D^{ij}_{e^ie^j}$, the sufficient condition obtained from Fig. \ref{fig:collision}(c) is
\begin{equation}
\label{BbelowDE}
    t^{ij,2}_{e^ie^j}\leq t^{ji,1}_{e^je^i}+M(1-\delta^{ij}_{e^ie^j}).
\end{equation}

% \begin{figure}[t]
% \centering
% \includegraphics[scale=0.475]{pictures/Case2.1.png}
% \caption{$t$-$s$ curves of Case 2.1.}
% \label{fig:Case2.1}
% \end{figure}

% \begin{figure}[t]
% \centering
% \includegraphics[scale=0.475]{pictures/Case2.2.png}
% \caption{$t$-$s$ curves of Case 2.2.}
% \label{fig:Case2.2}
% \end{figure}

\textit{Condition 2 of Case 1.2:} When $s^{ij,2}_{e^ie^j}>\hat{s}^{ji,1}_{e^je^i}+D^{ij}_{e^ie^j}$, the sufficient condition for collision avoidance is that the point B should stay below line CD and point D should stay above line AB (see Fig. \ref{fig:collision}(d)), which results in the following constraints:
\begin{equation}
\begin{aligned}
    t^{ij,2}_{e^ie^j}\leq&\frac{s^{ij,2}_{e^ie^j}-(\hat{s}^{ji,1}_{e^je^i}+D^{ij}_{e^ie^j})}{\hat{s}^{ji,2}_{e^je^i}-\hat{s}^{ji,1}_{e^je^i}}(t^{ji,2}_{e^je^i}-t^{ji,1}_{e^je^i}) + t^{ji,1}_{e^je^i}\\
    &+M(1-\delta^{ij}_{e^ie^j}),\\
\end{aligned}
\end{equation}
and
\begin{equation}
\begin{aligned}
    t^{ji,1}_{e^je^i}\geq&\frac{(\hat{s}^{ji,1}_{e^je^i}+D^{ij}_{e^ie^j})-s^{ij,1}_{e^ie^j}}{s^{ij,2}_{e^ie^j}-s^{ij,1}_{e^ie^j}}(t^{ij,2}_{e^ie^j}-t^{ij,1}_{e^ie^j})+t^{ij,1}_{e^ie^j}\\
    &-M(1-\delta^{ij}_{e^ie^j}).
\end{aligned}
\end{equation}

\textit{Case 2.1:} In this case, the $t$-$s$ curves of both CAVs are shown in Fig. \ref{fig:collision}(e). Similarly, we have the following results:
\begin{itemize}
    \item $s^{ij,1}_{e^ie^j}\in[\hat{s}^{ji,2}_{e^je^i}-D^{ij}_{e^ie^j}, \hat{s}^{ji,2}_{e^je^i}+D^{ij}_{e^ie^j}]$, and
    \item $s^{ij,2}_{e^ie^j}\in[\hat{s}^{ji,1}_{e^je^i}-D^{ij}_{e^ie^j}, \hat{s}^{ji,1}_{e^je^i}+D^{ij}_{e^ie^j}]$.
\end{itemize}

Noted that $s^{ij,1}_{e^ie^j}\in[\hat{s}^{ji,2}_{e^je^i}-D^{ij}_{e^ie^j}, \hat{s}^{ji,2}_{e^je^i}+D^{ij}_{e^ie^j}]$, the only possibility for collision avoidance is when (\ref{AaboveDE}) is satisfied. Also, since  $s^{ij,1}_{e^ie^j}\geq\hat{s}^{ji,2}_{e^je^i}-D^{ij}_{e^ie^j}$ always holds, \textit{Case 2.1} essentially coincides with \textit{Condition 2 of Case 1.1}.

\textit{Case 2.2:} As is shown in Fig. \ref{fig:collision}(f), since
$s^{ij,2}_{e^ie^j}\in[\hat{s}^{ji,1}_{e^je^i}-D^{ij}_{e^ie^j}, \hat{s}^{ji,1}_{e^je^i}+D^{ij}_{e^ie^j}]$, the only possibility for collision avoidance under this case is when (\ref{BbelowDE}) is satisfied. Also, since $s^{ij,2}_{e^ie^j}\leq\hat{s}^{ji,1}_{e^je^i}+D^{ij}_{e^ie^j}$ always holds, \textit{Case 2.2} essentially coincides with \textit{Condition 1 of Case 1.2}.

\begin{figure}[t]
\begin{align}\label{Collision2}
&\left\{
\begin{aligned}
    &(1-\theta^{ij,1}_{e^ie^j})t^i_{e^i_1}+\theta^{ij,1}_{e^ie^j} t^i_{e^i_2}+M(1-\delta^{ji}_{e^je^i})\\
    &\geq\frac{s^{ij,1}_{e^ie^j}-(\hat{s}^{ji,1}_{e^je^i}-D^{ij}_{e^ie^j})}{\hat{s}^{ji,2}_{e^je^i}-\hat{s}^{ji,1}_{e^je^i}}
    ((1-\theta^{ji,2}_{e^je^i})t^j_{e^j_1}+\theta^{ji,2}_{e^je^i} t^j_{e^j_2})\\
    &+(1-\frac{s^{ij,1}_{e^ie^j}-(\hat{s}^{ji,1}_{e^je^i}-D^{ij}_{e^ie^j})}{\hat{s}^{ji,2}_{e^je^i}-\hat{s}^{ji,1}_{e^je^i}})((1-\theta^{ji,1}_{e^je^i})t^j_{e^j_1}+\theta^{ji,1}_{e^je^i} t^j_{e^j_2}),\\
    &(1-\theta^{ji,2}_{e^je^i})t^j_{e^j_1}+\theta^{ji,2}_{e^je^i} t^j_{e^j_2}-M(1-\delta^{ji}_{e^je^i})\\
    &\leq\frac{(\hat{s}^{ji,2}_{e^je^i}-D^{ij}_{e^ie^j})-s^{ij,1}_{e^ie^j}}{s^{ij,2}_{e^ie^j}-s^{ij,1}_{e^ie^j}}((1-\theta^{ij,2}_{e^ie^j})t^i_{e^i_1}+\theta^{ij,2}_{e^ie^j}t^i_{e^i_2})\\
    &+(1-\frac{(\hat{s}^{ji,2}_{e^je^i}-D^{ij}_{e^ie^j})-s^{ij,1}_{e^ie^j}}{s^{ij,2}_{e^ie^j}-s^{ij,1}_{e^ie^j}})((1-\theta^{ij,1}_{e^ie^j})t^i_{e^i_1}+\theta^{ij,1}_{e^ie^j}t^i_{e^i_2})\\
    &\ \ \ \ \ \ \ \ \ \ \ \ \ \ \ \ \ \ \ \ \ \ \ \ \ \ \ \ \ \ \ \ \ \ \ \ \ \ \ \ \ \ \textup{if}\ s^{ij,1}_{e^ie^j}<\hat{s}^{ji,2}_{e^je^i}-D^{ij}_{e^ie^j}\\
    &(1-\theta^{ij,1}_{e^ie^j})t^i_{e^i_1}+\theta^{ij,1}_{e^ie^j} t^i_{e^i_2}\\
    &\geq (1-\theta^{ji,2}_{e^je^i})t^j_{e^j_1}+\theta^{ji,2}_{e^je^i} t^j_{e^j_2}-M(1-\delta^{ji}_{e^je^i})\\
    &\ \ \ \ \ \ \ \ \ \ \ \ \ \ \ \ \ \ \ \ \ \ \ \ \ \ \ \ \ \ \ \ \ \ \ \ \ \ \ \ \ \ \ \ \ \ \ \ \ \ \ \ \ \ \ \ \ \ \ \ \ \ \ \ \ \textup{else}
\end{aligned}
\right.\nonumber\\
&\forall \{(i,e^i),(j,e^j)\}\in\Gamma.
\end{align}
\end{figure}

\begin{figure}[t]
\begin{align}\label{Collision3}
&\left\{
\begin{aligned}
    &(1-\theta^{ij,2}_{e^ie^j})t^i_{e^i_1}+\theta^{ij,2}_{e^ie^j} t^i_{e^i_2}-M(1-\delta^{ij}_{e^ie^j})\\
    &\leq\frac{s^{ij,2}_{e^ie^j}-(\hat{s}^{ji,1}_{e^je^i}+D^{ij}_{e^ie^j})}{\hat{s}^{ji,2}_{e^je^i}-\hat{s}^{ji,1}_{e^je^i}}((1-\theta^{ji,2}_{e^je^i})t^j_{e^j_1}+\theta^{ji,2}_{e^je^i} t^j_{e^j_2})\\
    &+(1-\frac{s^{ij,2}_{e^ie^j}-(\hat{s}^{ji,1}_{e^je^i}+D^{ij}_{e^ie^j})}{\hat{s}^{ji,2}_{e^je^i}-\hat{s}^{ji,1}_{e^je^i}})((1-\theta^{ji,1}_{e^je^i})t^j_{e^j_1}+\theta^{ji,1}_{e^je^i} t^j_{e^j_2}),\\
    &(1-\theta^{ji,1}_{e^je^i})t^j_{e^j_1}+\theta^{ji,1}_{e^je^i} t^j_{e^j_2}+M(1-\delta^{ij}_{e^ie^j})\\
    &\geq\frac{(\hat{s}^{ji,1}_{e^je^i}+D^{ij}_{e^ie^j})-s^{ij,1}_{e^ie^j}}{s^{ij,2}_{e^ie^j}-s^{ij,1}_{e^ie^j}}((1-\theta^{ij,2}_{e^ie^j})t^i_{e^i_1}+\theta^{ij,2}_{e^ie^j} t^i_{e^i_2})\\
    &+(1-\frac{(\hat{s}^{ji,1}_{e^je^i}+D^{ij}_{e^ie^j})-s^{ij,1}_{e^ie^j}}{s^{ij,2}_{e^ie^j}-s^{ij,1}_{e^ie^j}})((1-\theta^{ij,1}_{e^ie^j})t^i_{e^i_1}+\theta^{ij,1}_{e^ie^j} t^i_{e^i_2})\\
    &\ \ \ \ \ \ \ \ \ \ \ \ \ \ \ \ \ \ \ \ \ \ \ \ \ \ \ \ \ \ \ \ \ \ \ \ \ \ \ \ \ \ \textup{if}\ s^{ij,2}_{e^ie^j}>\hat{s}^{ji,1}_{e^je^i}+D^{ij}_{e^ie^j}\\
    &(1-\theta^{ij,2}_{e^ie^j})t^i_{e^i_1}+\theta^{ij,2}_{e^ie^j} t^i_{e^i_2}\\
    &\leq (1-\theta^{ji,1}_{e^je^i})t^j_{e^j_1}+\theta^{ji,1}_{e^je^i} t^j_{e^j_2}+M(1-\delta^{ij}_{e^ie^j}).\\
    &\ \ \ \ \ \ \ \ \ \ \ \ \ \ \ \ \ \ \ \ \ \ \ \ \ \ \ \ \ \ \ \ \ \ \ \ \ \ \ \ \ \ \ \ \ \ \ \ \ \ \ \ \ \ \ \ \ \ \ \ \ \ \ \ \ \textup{else}\\
\end{aligned}
\right.\nonumber\\
&\forall \{(i,e^i),(j,e^j)\}\in\Gamma.
\end{align}
\end{figure}

Gathering all the cases discussed above and plugging in (\ref{linearInterpolation}), we obtain the collision avoidance constraints as in (\ref{Collision2}) and (\ref{Collision3}), both of which are linear with respect to time $t^i_{e^i_1}$, $t^i_{e^i_2}$, $t^j_{e^j_1}$, and $t^j_{e^j_2}$.

\subsection{MILP Formulation}
To facilitate cooperative passing through complicated traffic scenarios while improving overall traffic efficiency and comfort of passengers, we consider important performance indices including total travel time, velocities, accelerations, and steering effects. 

The objective function corresponding to total travel time is given as
\begin{equation}
    f_t = \sum_{i\in\mathcal{N}}\sum_{v\in des^i}t^i_v.
\end{equation}
The objective function corresponding to the velocities of vehicles is given as
\begin{equation}
    f_V = \sum_{i\in\mathcal{N}}\sum_{e\in\mathcal{E}^i}(\Bar{s}^i_e+\underline{s}^i_e).
\end{equation}
By including $f_V$ as a penalty term, CAVs are urged to track the reference velocities as closely as possible. The objective function corresponding to the accelerations of vehicles is given as
\begin{align}
    f_a &= \sum_{i\in\mathcal{N}}\sum_{v\in\Bar{\mathcal{V}}^i}\sum_{\alpha\in\mathcal{E}^{i,in}_v}\sum_{\beta\in\mathcal{E}^{i,out}_v}\sum_{k\in\mathcal{K}}(V^{i,k}_r)^2(\Bar{\gamma}^{i,k}_{v,\alpha,\beta}+\underline{\gamma}^{i,k}_{v,\alpha,\beta})\nonumber\\
    &+\sum_{i\in\mathcal{N}}\sum_{\beta\in\mathcal{E}^{i,out}_{s^i}}\sum_{k\in\mathcal{K}}(V^{i,k}_r)^2(\Bar{\gamma}^{i,k}_{s^i,\beta}+\underline{\gamma}^{i,k}_{s^i,\beta}).
\end{align}
By including $f_a$ as a penalty term, longitudinal accelerations and decelerations are penalized to improve the passenger comfort. The objective function corresponding to the steering angle is given as
\begin{equation}
\begin{aligned}
f_\theta&=\sum_{i\in\mathcal{N}}\sum_{v\in\Bar{\mathcal{V}}^i}\sum_{\alpha\in\mathcal{E}^{i,in}_v}\sum_{\beta\in\mathcal{E}^{i,out}_v}\sum_{k\in\mathcal{K}}\eta^{i,k}_{v,\alpha,\beta}\\
&+\sum_{i\in\mathcal{N}}\sum_{\beta\in\mathcal{E}^{i,out}_{s^i}}\sum_{k\in\mathcal{K}}\eta^{i,k}_{s^i,\beta}.
\end{aligned}
\end{equation}
By including $f_\theta$ as a penalty term, lateral accelerations induced by steering are penalized, such that the proposed method will not generate zigzag paths and sharp turns for CAVs to follow.

With the above discussions, we introduce the following optimization problem:
\begin{equation}
\label{MILP}
\begin{aligned}
    \min\ &\alpha_tf_t+\alpha_Vf_V+\alpha_af_a+\alpha_\theta f_\theta\\
    s.t.\ &(\ref{start})\text{-}(\ref{continuity}),(\ref{velocity}),(\ref{velocitySlack}),(\ref{identifyRegion}),(\ref{acceleration}),(\ref{initIdentifyRegion}),(\ref{initAcceleration}),\\
    &(\ref{steering}),(\ref{initSteering}),(\ref{Collision1}),(\ref{Collision2}),(\ref{Collision3})
\end{aligned}
\end{equation}
where $\alpha_t$, $\alpha_V$, $\alpha_a$, and $\alpha_\theta$ are weighting parameters corresponding to each of the objectives. Since all the objectives and constraints are linear with respect to optimization variables, problem (\ref{MILP}) is a standard MILP problem that can be solved by off-the-shelf solvers.

\subsection{Trajectory Planning}
The paths and the associated time profiles obtained from the previous section can serve as good references for generating trajectories that are feasible and collision-free for all CAVs. In this section, we utilize the paths and the time profiles obtained by solving (\ref{MILP}) as references and formulate the cooperative trajectory planning problem as an optimal control problem. Kinematic constraints and other pertinent constraints are properly considered and addressed.

\subsubsection{Kinematic and Physical Constraints}
For all CAVs, we adopt the discrete-time bicycle kinematic model with constant time interval $\tau_s$. We assume that the initial time stamps for all CAVs are aligned with each other. Although the durations of the generated references for CAVs by solving (\ref{MILP}) are different from each other and so as the resulting trajectories, we can cut the length of all trajectories to the shortest one to ensure that their durations are the same. As a result, the initial time stamps of all CAVs for the next planning period will still be aligned. We denote the reference path for CAV $i$ obtained by solving (\ref{MILP})
as $\{v^{i,k}\}$, where $v^{i,k}$ is the $k\text{-}th$ vertex passed through by CAV $i$ in the path and $k\in\{0,1,...,n^i\}$. For simplicity, we denote the desired time for the center of CAV $i$ to reach $v^{i,k}$ as $t^{i,k}$, and the corresponding location of $v^{i,k}$ as $p^{i,k}$. We further assume that $t^{i,0}=0$ for all $i\in\mathcal{N}$ and the first time stamp is $0$. Then the last time stamp for CAV $i$ is $\tau^i=\textup{int}(t^{i,n^i}/\tau_s)$. The set of all time stamps for CAV $i$ is then denoted as $T^i=\{0,1,...,\tau^i\}$.

In particular, the state vector for CAV $i$ at the time stamp $\tau\in T^i$, $x^i_{\tau}$, is defined as $x^i_{\tau}=(p^i_{x,\tau},p^i_{y,\tau},\theta^i_{\tau},v^i_{\tau})$. $p^i_{x,\tau}$ and $p^i_{y,\tau}$ are the X and Y coordinates of the center of the rear axle of CAV $i$ in the global coordinate, $\theta^i_{\tau}$ is the heading angle, and $v^i_{\tau}$ denotes the longitudinal velocity. Meanwhile, the control input vector $u^i_{\tau}$ is given as $u^i_{\tau}=(\delta^i_{\tau},a^i_{\tau})$, where $\delta^i_{\tau}$ is the steering angle and $a^i_{\tau}$ is the acceleration. With the above definitions, the following equations characterize the discrete-time bicycle kinematic model~\cite{tassa2014control}:
\begin{equation}
\label{dynamics}
\left\{
\begin{aligned}
p^i_{x,\tau+1} &= p^i_{x,\tau}+f_r(v^i_{\tau},\delta^i_{\tau})\cos(\theta^i_{\tau}),\\
p^i_{y,\tau+1} &= p^i_{y,\tau}+f_r(v^i_{\tau},\delta^i_{\tau})\sin(\theta^i_{\tau}),\\
\theta^i_{\tau+1} &= \theta^i_{\tau}+\arcsin\left(\tau_s v^i_{\tau}\sin(\delta^i_{\tau})/b^i\right),\\
v^i_{\tau+1} &= v^i_{\tau}+\tau_sa^i_{\tau}.
\end{aligned}
\right.
\end{equation}
In particular, $b^i$ is the wheelbase of CAV $i$ and $\tau_s$ is the time interval. Moreover, the function $f_r(v^i_{\tau},\delta^i_{\tau})$ is defined as
\begin{equation}
    f_r(v,\delta) = b+\tau_sv\cos(\delta)-\sqrt{b^2-(\tau_sv\sin(\delta))^2}.
\end{equation}
The above kinematic constraints can be equivalently expressed in a short form as
\begin{equation}
\label{kinematics}
    x^i_{\tau+1}=f(x^i_{\tau},u^i_{\tau}), \forall i\in\mathcal{N}, \forall \tau \in \{0,1,...,\tau^i-1\}.
\end{equation}

Meanwhile, due to physical limitations on engine forces, break forces, and wheel angles, the following constraints need to be imposed on the control inputs:
\begin{equation}
\label{controlLimits}
\begin{aligned}
&\begin{aligned}
    a_\textup{min}&\leq a^i_\tau\leq a_\textup{max},\\
    \delta_\textup{min}&\leq\delta^i_\tau\leq\delta_\textup{max},
\end{aligned}\\
&\forall i\in\mathcal{N}, \forall \tau \in \{0,1,...,\tau^i-1\}.
\end{aligned}
\end{equation}

\subsubsection{Collision Avoidance Constraints}
Although conflicts between CAVs are resolved to a large extent by enforcing (\ref{Collision1}), (\ref{Collision2}), and (\ref{Collision3}) in solving for the optimal solution in problem (\ref{MILP}), slight collisions may still occur when full vehicle kinematic systems are involved and thus the real trajectories for CAVs may deviate from the reference paths. To avoid possible collisions strictly, relevant constraints must be imposed on the optimal control problem. We assume that all CAVs are of the same size, and we represent each of the CAVs with two identical circles aligned in the longitudinal direction. For CAV $i$, we denote the positions of centers of the front circle and the rear circle at the time stamp $\tau\in T^i$ as $p^i_{f,\tau}$ and $p^i_{r,\tau}$. Sufficient conditions for collision avoidance are that the distance between any pairs of circles for different CAVs should be greater than the safety distance, namely
\begin{equation}
\label{fullCollision}
\begin{aligned}
    &||p^i_{\beta,\tau}-p^j_{\gamma,\tau}||_2\geq d_\textup{safe},\\
    &\forall i,j\in\mathcal{N}, i\neq j, \forall\tau\in T^i\cap T^j, \forall \beta,\gamma\in\{f,r\}.
\end{aligned}
\end{equation}

\subsubsection{Reference Paths}
Linear interpolation is used to generate the reference point for each time stamp. In particular, the corresponding reference position is
\begin{equation}
\begin{aligned}
&\hat{p}^i_{\tau,ref} = \frac{\tau\tau_s-t^{i,k}}{t^{i,k+1}-t^{i,k}}p^{i,k+1}+\frac{t^{i,k+1}-\tau\tau_s}{t^{i,k+1}-t^{i,k}}p^{i,k},\\
&k\in\{0,1,...,n^i\},t^{i,k}\leq\tau\tau_s,t^{i,k+1}\geq\tau\tau_s,\\
&\forall i\in\mathcal{N}, \forall \tau \in T^i.
\end{aligned}
\end{equation}

Meanwhile, $\hat{p}^i_{\tau,ref}$ is the reference position of the center of the CAV $i$. However, due to the characteristics of the vehicle kinematics, optimization is performed over the midpoint of the rear axle. To bridge the gap, we derive the corresponding reference position for the midpoint of the rear axle as
\begin{equation}
\begin{aligned}
&p^i_{\tau,ref} = \hat{p}^i_{\tau,ref}-d^i_b\frac{p^{i,k+1}-p^{i,k}}{||p^{i,k+1}-p^{i,k}||_2},\\
&k\in\{0,1,...,n^i\},t^{i,k}\leq\tau\tau_s,t^{i,k+1}\geq\tau\tau_s.\\
&\forall i\in\mathcal{N}, \forall \tau \in T^i,
\end{aligned}
\end{equation}
where $d^i_b$ is the distance between the center of CAV $i$ and the midpoint of the rear axle, and $||\cdot||_2$ denotes the $L_2$-norm.

\begin{figure*}[t]

\subfigure[Waypoint graph]{\includegraphics[scale=0.66]{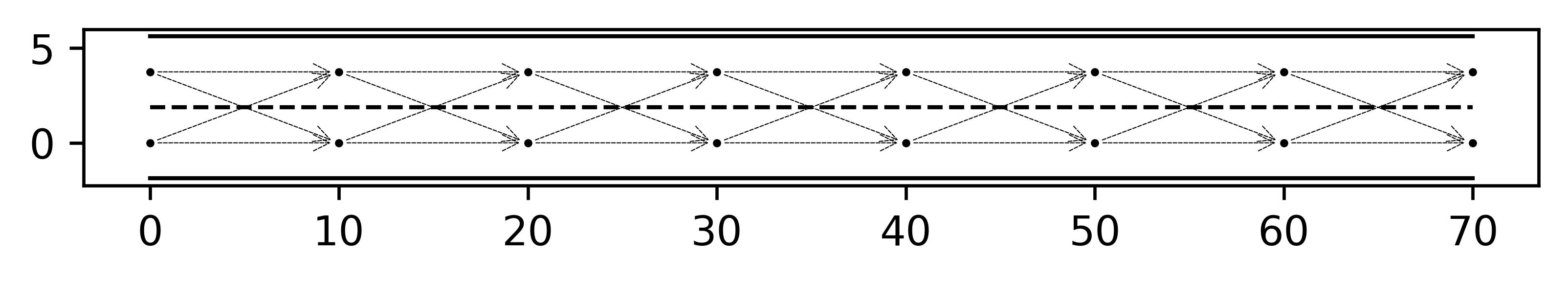}}
\centering
\subfigure[$t=0\,\textup{s}$]{\includegraphics[scale=0.66]{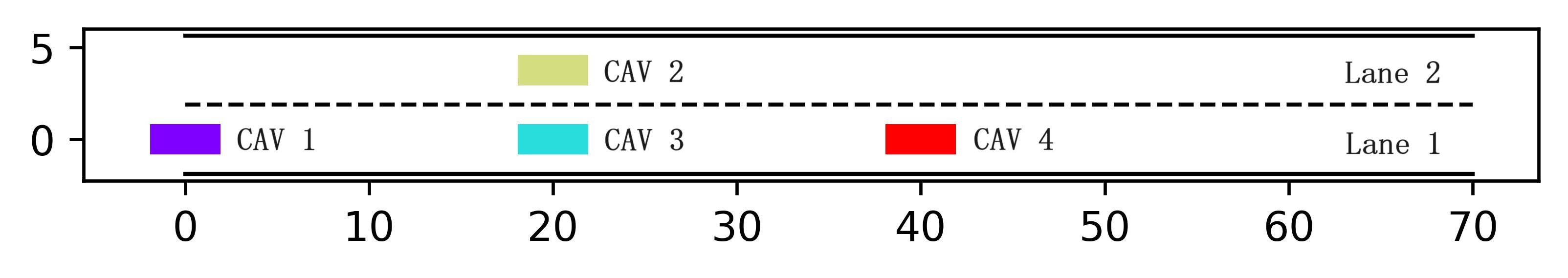}}
\subfigure[$t=1.5\,\textup{s}$]{\includegraphics[scale=0.66]{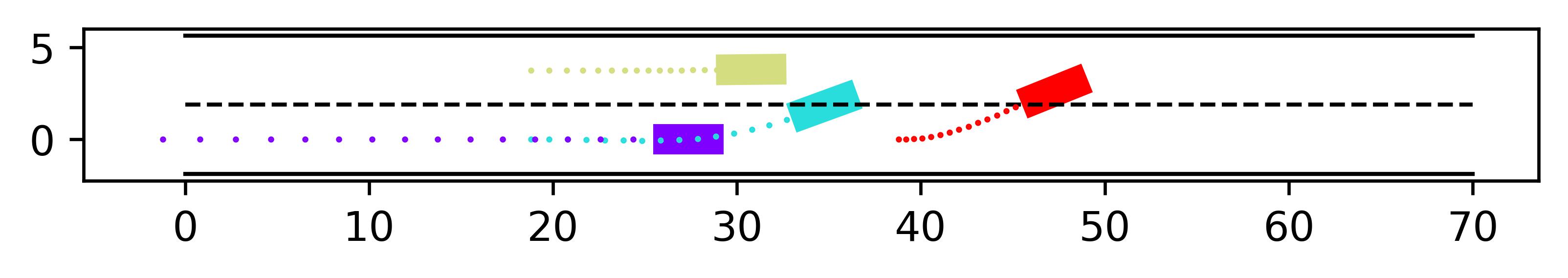}}
\subfigure[$t=2.3\,\textup{s}$]{\includegraphics[scale=0.66]{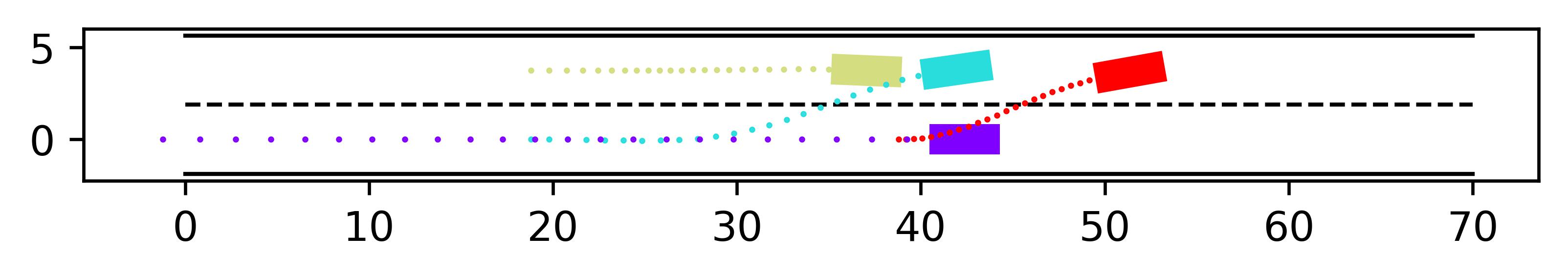}}
\subfigure[$t=3.4\,\textup{s}$]{\includegraphics[scale=0.66]{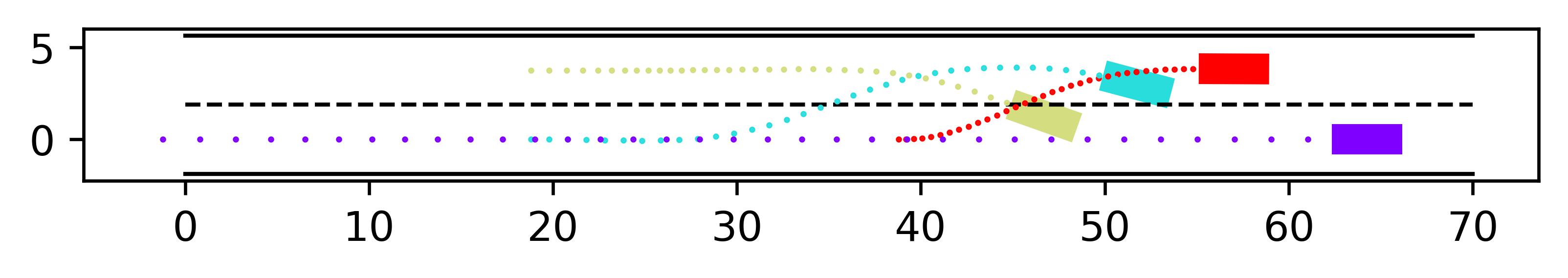}}
\subfigure[$t=4.5\,\textup{s}$]{\includegraphics[scale=0.66]{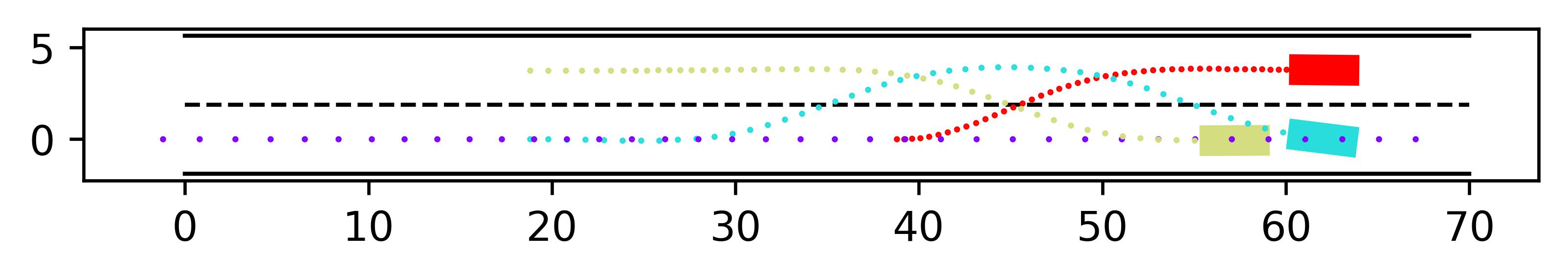}}

\caption{Simulation results for the overtaking scenario.}

\label{fig:overtaking}
\end{figure*}

\subsubsection{Problem Formulation}
In the optimal control problem, we minimize both the control inputs and the deviations of actual trajectories from the reference, namely
\begin{equation}
\label{OCP}
\begin{aligned}
    \min\limits_{\{u^i_\tau\}}&\ \sum^N_{i=1}\left(\sum_{\tau=0}^{\tau^i}||x^i_\tau-x^i_{\tau,ref}||^2_Q+\sum^{\tau^i-1}_{\tau=0}||u^i_\tau||^2_R\right)\\
    \textup{s.t.}&\ (\ref{kinematics}),(\ref{controlLimits}),(\ref{fullCollision}).
\end{aligned}
\end{equation}

In particular, $x^i_{\tau,ref}=(p^i_{x,\tau,ref},p^i_{y,\tau,ref},\theta^i_{\tau,ref},v^i_{\tau,ref})$, where $p^i_{x,\tau,ref}$ and $p^i_{y,\tau,ref}$ are the global X and Y coordinates of reference position $p^i_{\tau,ref}$ respectively, $\theta^i_{\tau,ref}$ is the reference heading angle, and $v^i_{\tau,ref}$ is the reference velocity. $||\boldsymbol{\cdot}||_Q$ and $||\boldsymbol{\cdot}||_R$ denote the $L_2$-norm weighted by matrices $Q$ and $R$, where $Q$ is positive semi-definite and $R$ is positive definite. Solving (\ref{OCP}) yields feasible and collision-free trajectories for all CAVs involved.

\section{Simulation Results}

To verify the effectiveness of the proposed method, case studies in three different traffic scenarios are presented and discussed in this section. Gurobi~\cite{gurobi} and CasADi~\cite{andersson2019casadi} are used to solve problems (\ref{MILP}) and (\ref{OCP}), respectively. Pertinent parameter settings are presented in Table I. In particular, $b$, $L$, and $W$ are the wheelbase, length, and width for all CAVs, respectively. $d_f$ and $d_r$ are the distances from the center of the rear axle to the front and the rear circle, respectively. Moreover, the penalty matrices in problem (\ref{OCP}) are set as $Q=\textup{diag}(20,20,0,0)$ and $R=\textup{diag}(20,0.1)$.

\begin{table}[t]
\centering
\caption{Parameter settings used in simulations}
\begin{tabular}{p{0.7cm}p{1.3cm}p{0.7cm}p{1.3cm}p{0.7cm}p{1.0cm}}
\hline
Param.               & Value                            & Param.              & Value 
& Param.              & Value \\ \hline

$\gamma_\textup{max}$                      & 3.0\,$\textup{m}/\textup{s}^2$                        & $\gamma_\textup{min}$              & -4.5\,$\textup{m}/\textup{s}^2$
&$\eta_\textup{max} $                & 3.0\,$\textup{m}/\textup{s}^2$  \\

$\alpha_t $       & 0.1 &  $\alpha_V$                 & 1.0 & $\alpha_a$                   & 0.5  \\

$\alpha_\theta$         & 0.5 & $\tau_s$   & 0.1\,\textup{s}  
&  $b$             & 2.405\,m \\                     

$d_\textup{safe}$         & 2.366\,m                      &$L$    & 3.826\,m                      & $W$               & 1.673\,m   \\

$d_f$        & 2.279\,m                          & $d_r$                    & 0.126\,m  
&$a_\textup{max}$                   & 4.0\,$\textup{m}/\textup{s}^2$   \\   

$a_\textup{min}$                    & -6.0\,$\textup{m}/\textup{s}^2$  & $\delta_\textup{max}$                        & 0.6\,rad                       & $\delta_\textup{min}$    & -0.6\,rad  \\ \hline
\end{tabular}
\end{table}

\subsection{Overtaking}

\begin{figure}[t]
\centering
\includegraphics[scale=0.60]{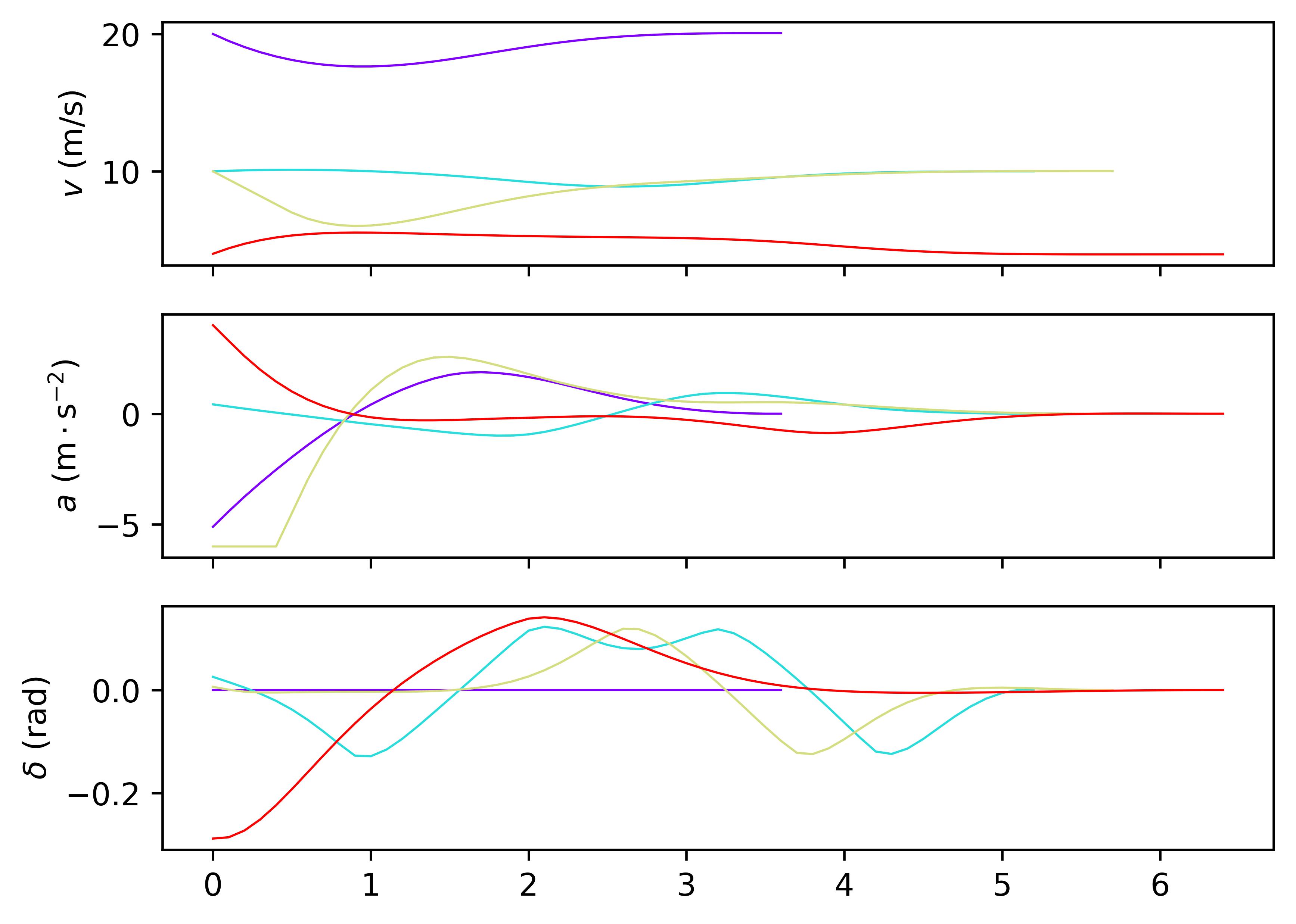}
\caption{Velocities, accelerations, and steering angles for all CAVs in the overtaking scenario.}
\label{fig:v-t-S}
\end{figure}

\begin{figure*}[t]

\subfigure[Waypoint graph]{\includegraphics[scale=0.66]{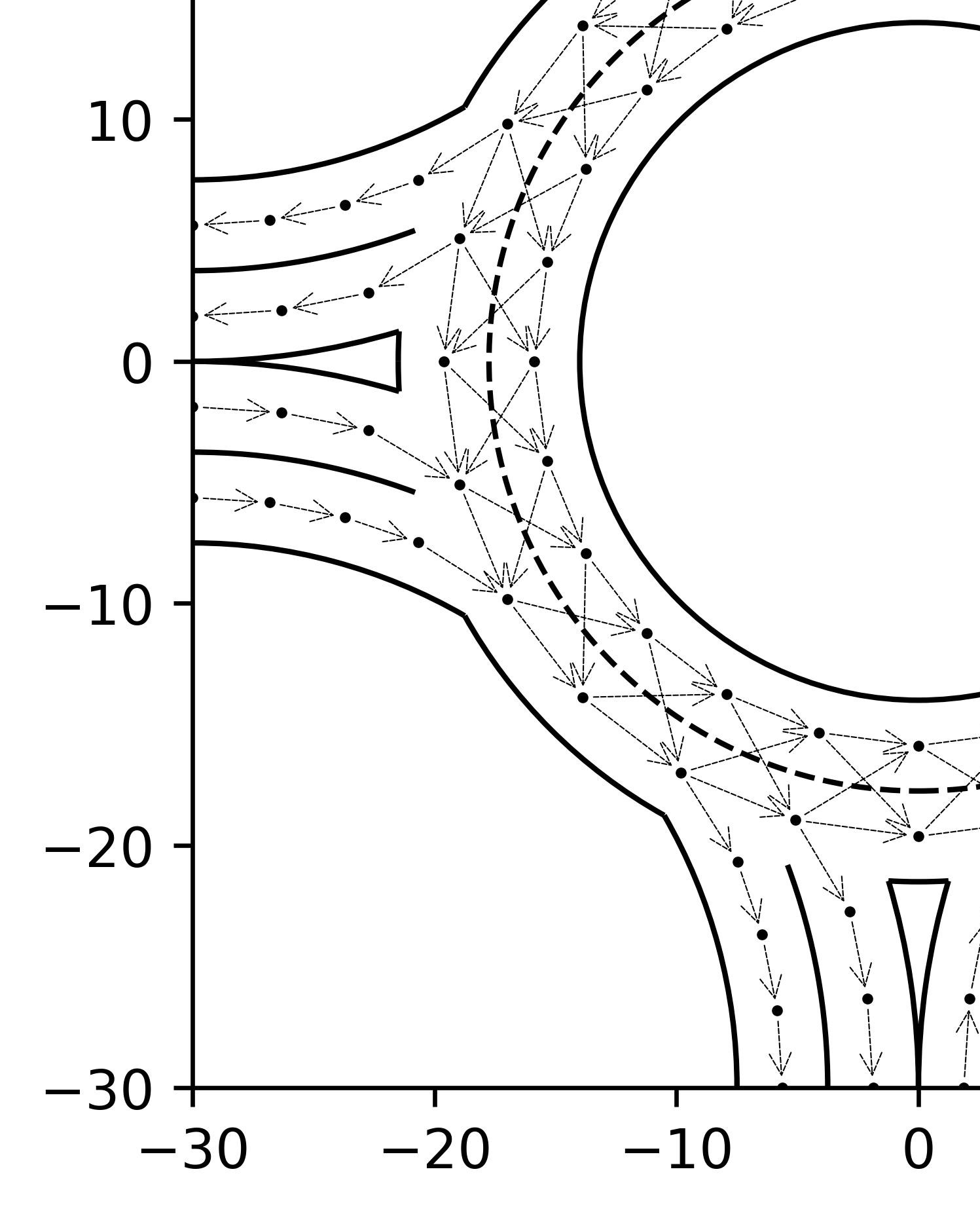}}
\subfigure[$t=0\,\textup{s}$]{\includegraphics[scale=0.66]{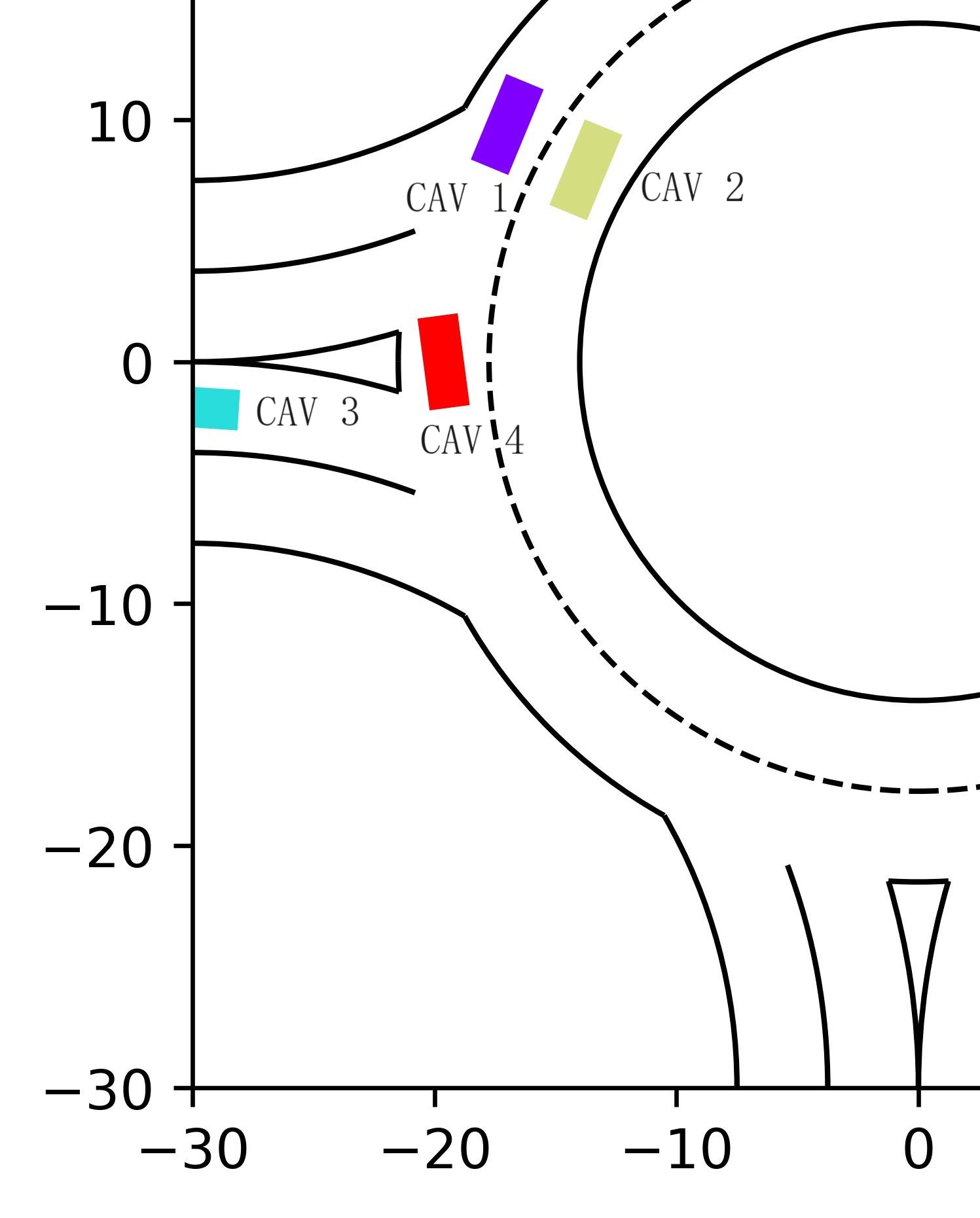}}
\subfigure[$t=1.0\,\textup{s}$]{\includegraphics[scale=0.66]{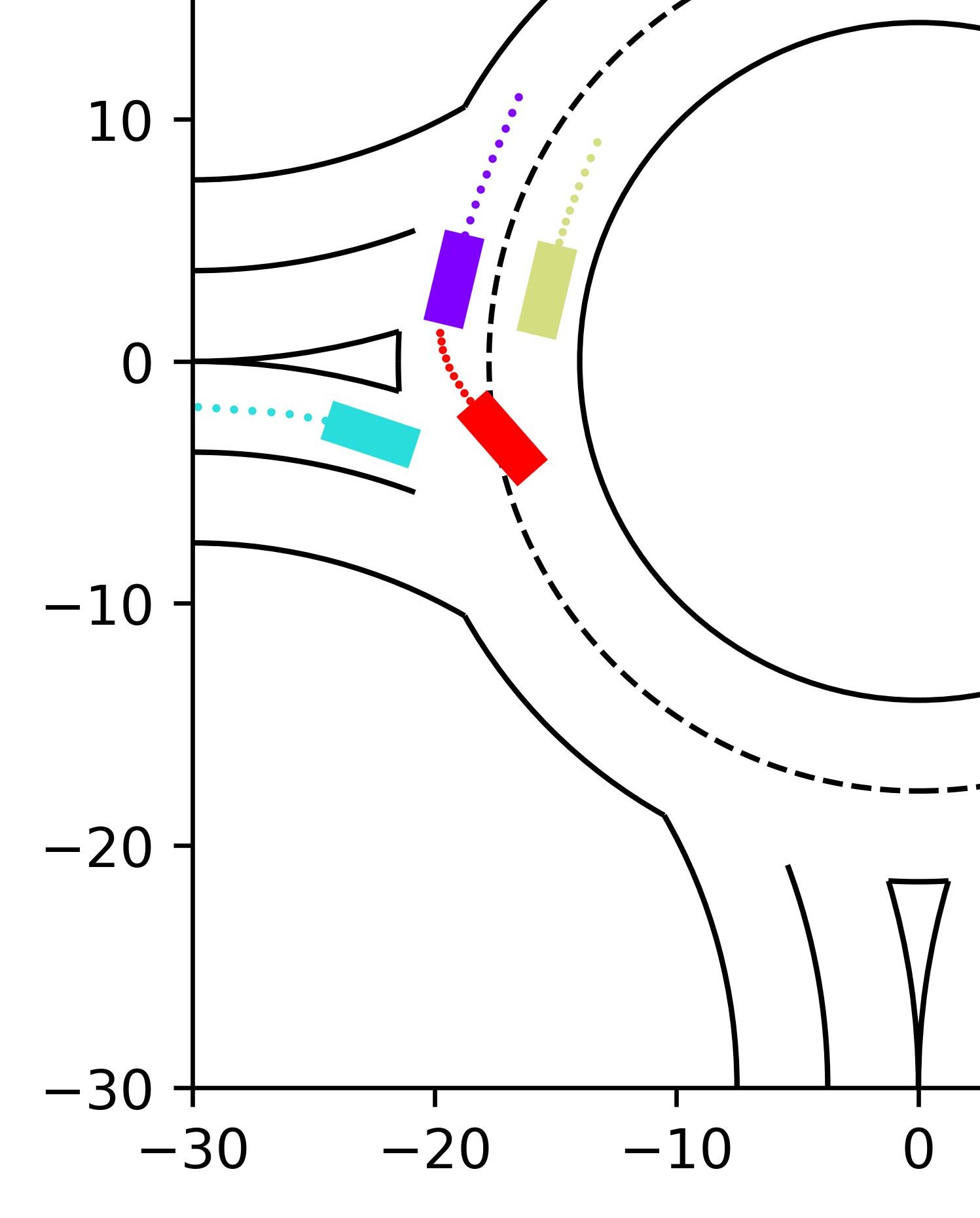}}
\subfigure[$t=2.0\,\textup{s}$]{\includegraphics[scale=0.66]{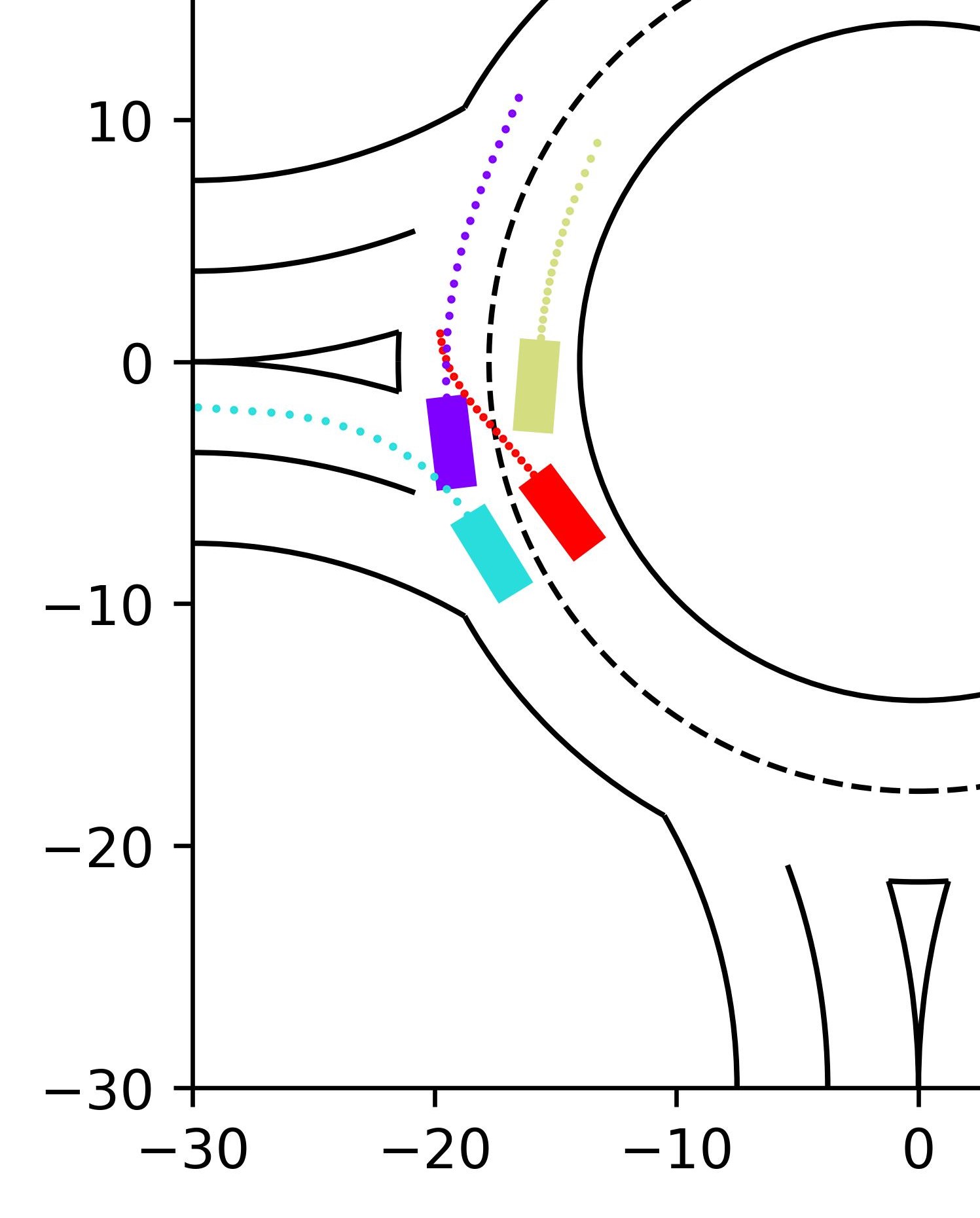}}
\centering
\subfigure[$t=3.0\,\textup{s}$]{\includegraphics[scale=0.66]{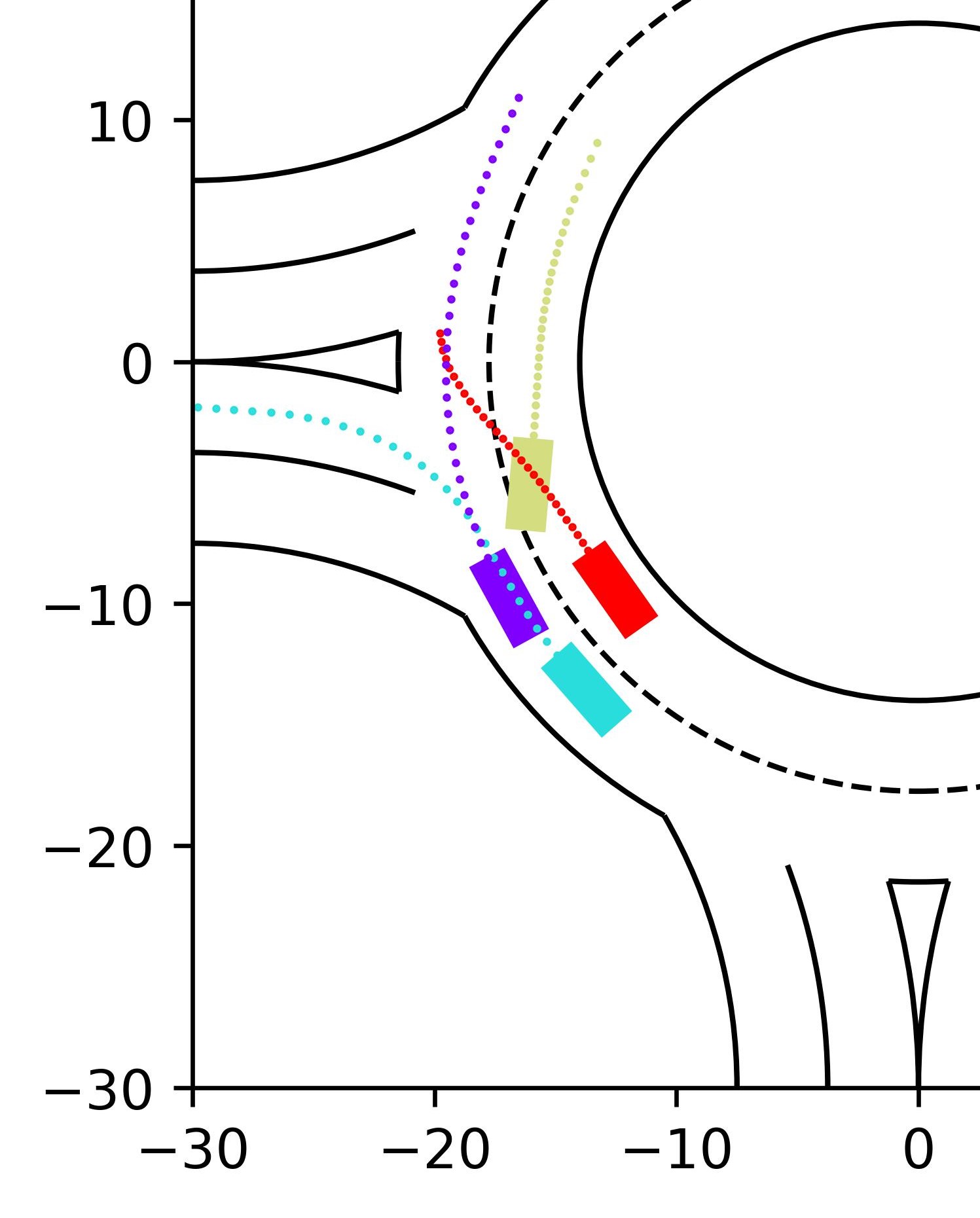}}
\subfigure[$t=4.0\,\textup{s}$]{\includegraphics[scale=0.66]{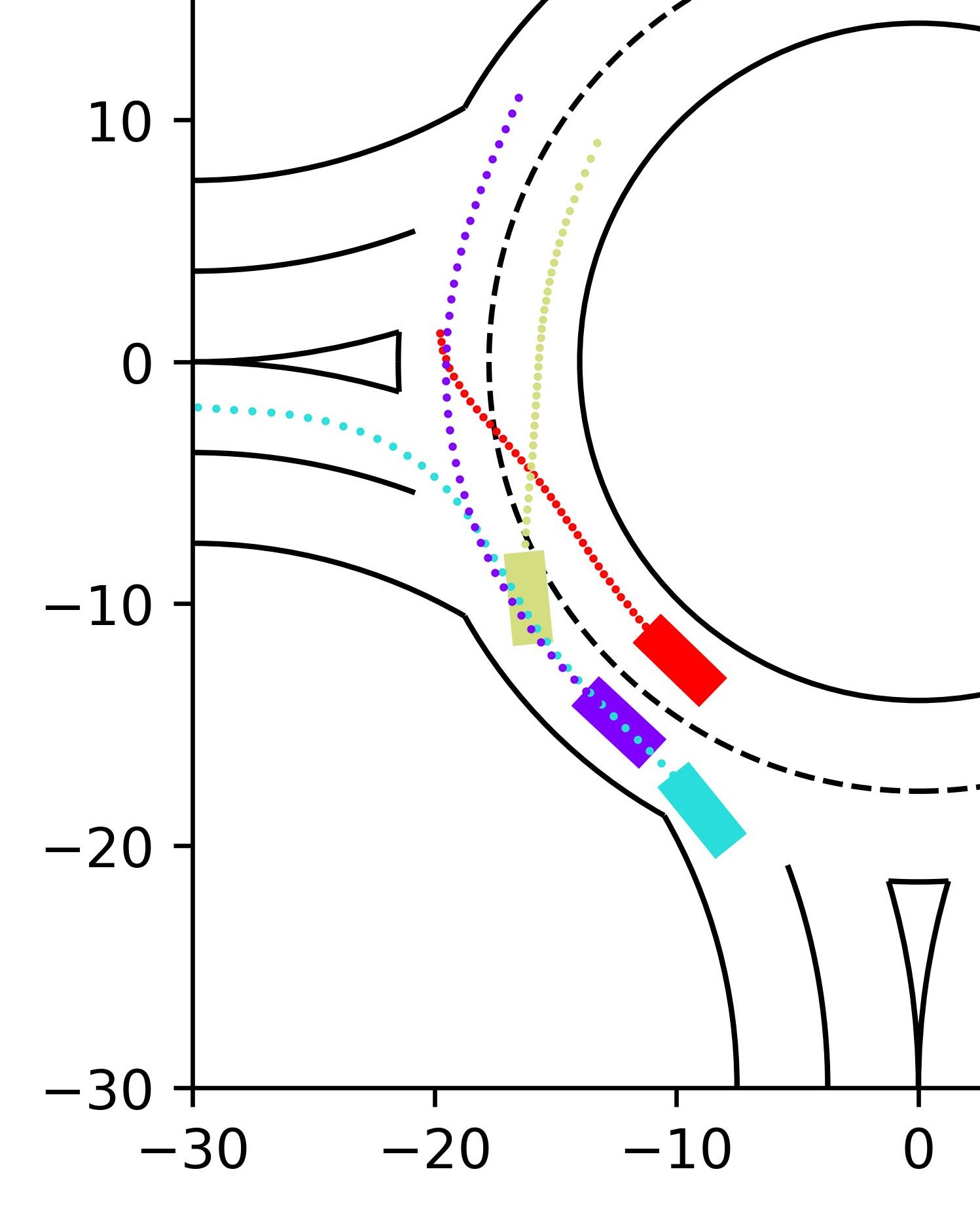}}
\subfigure[$t=5.0\,\textup{s}$]{\includegraphics[scale=0.66]{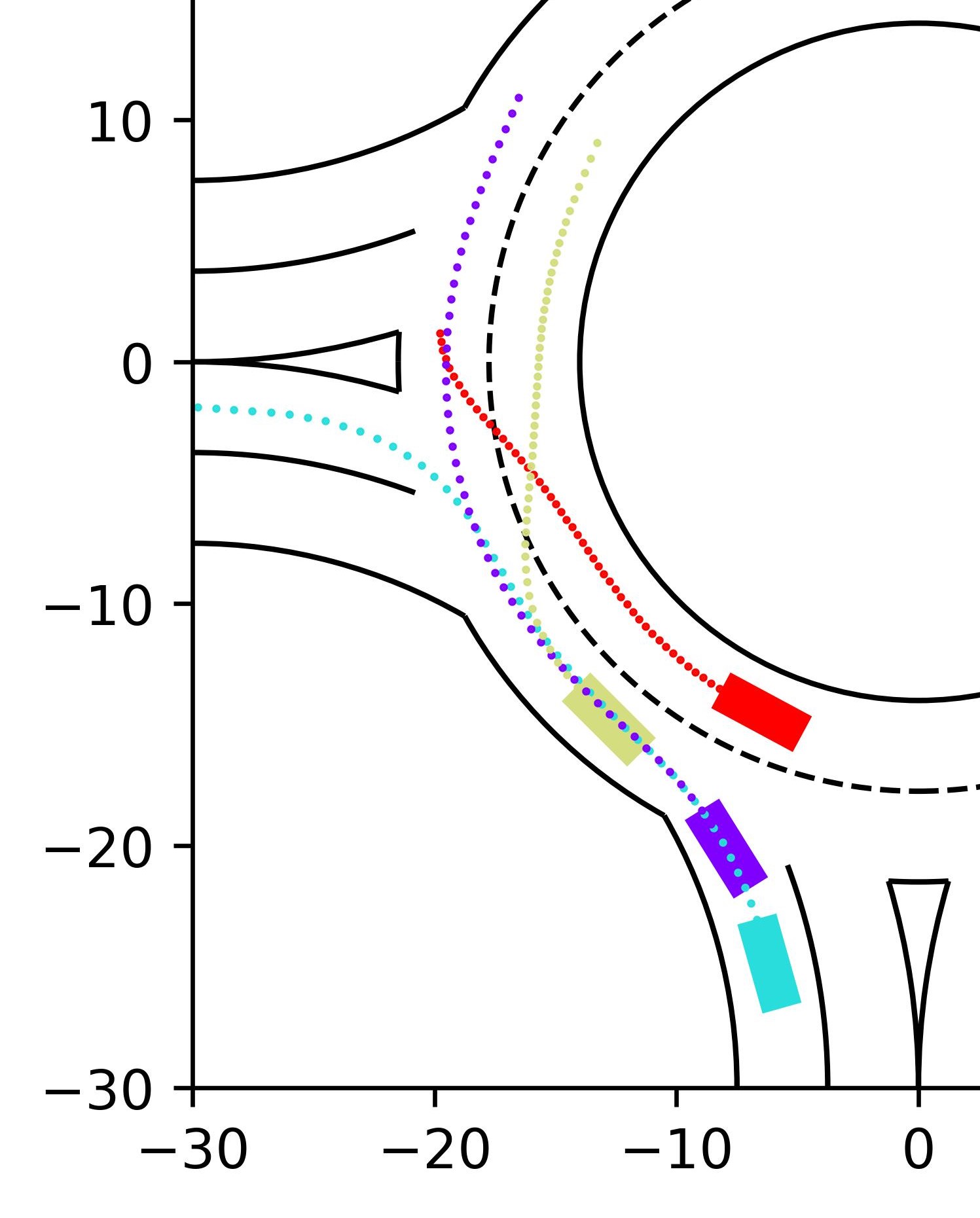}}
\subfigure[$t=6.0\,\textup{s}$]{\includegraphics[scale=0.66]{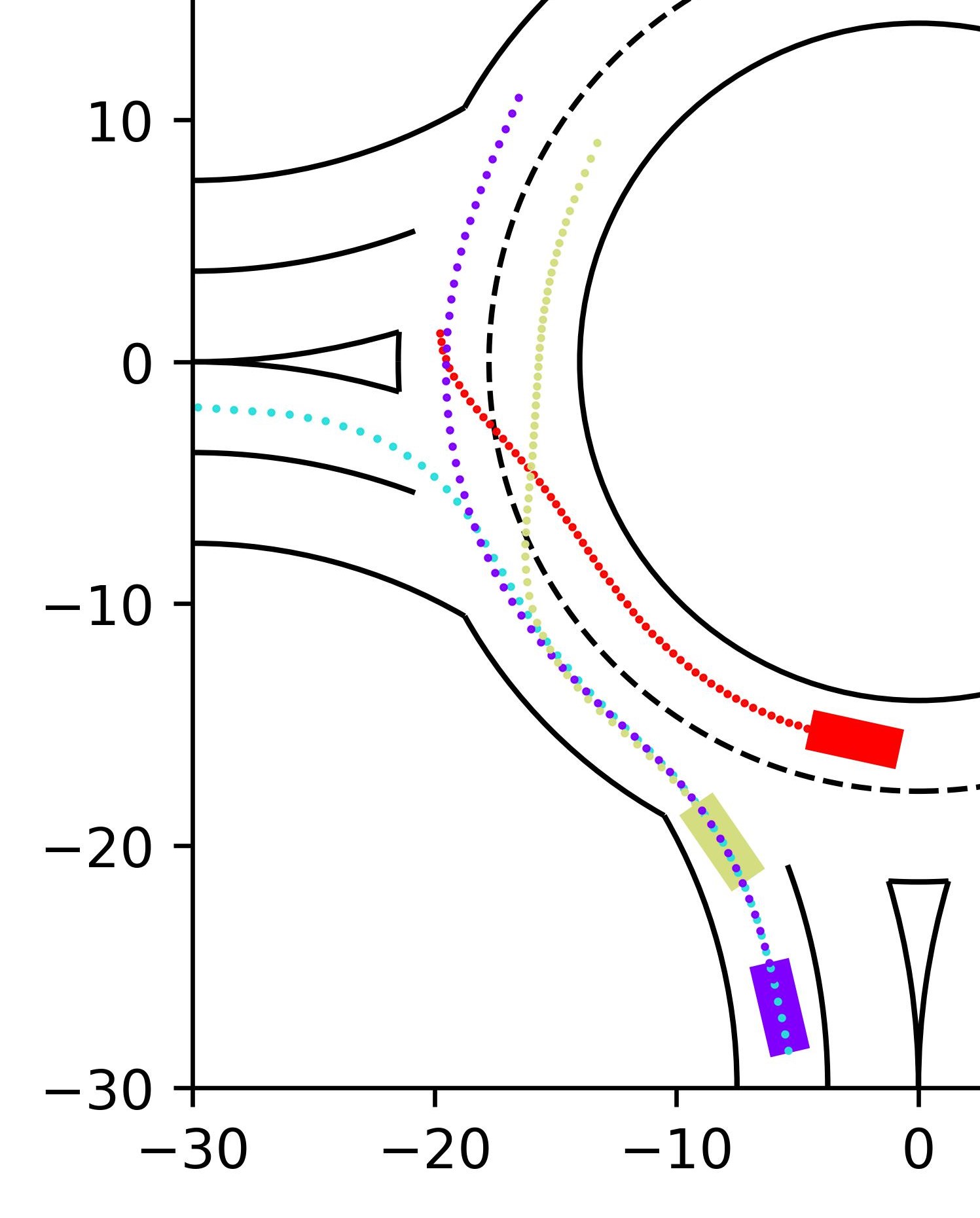}}

\caption{Simulation results for the roundabout scenario.}

\label{fig:roundabout}
\end{figure*}

In this section, we present an overtaking example for CAVs on a two-lane, one-way road. The waypoint graph corresponding to this traffic scenario is presented in Fig. \ref{fig:overtaking}(a). For the road section that is 70 meters in length, we sample along the center line of both lanes with a sampling interval of 10 meters. Each of the sampled waypoints is then connected to the next waypoint on the same lane and also the next waypoint on the other lane to constitute the set of edges. The width of each lane is set to be 3.75\,m. The initial positions of the four CAVs involved in this traffic scenario are given in Fig. \ref{fig:overtaking}(b). In particular, the initial velocities for CAV 1-4 are 20\,m/s, 10\,m/s, 10\,m/s, and 4\,m/s, respectively, and the reference velocities are set to be the same as the initial velocities. $V^i_{fast}$ and $V^i_{slow}$ are set to be $1.3\,V^i_r$ and $0.6\,V^i_r$ for all CAVs, respectively. Obviously, the CAVs behind are running at much higher velocities than the CAVs in the front. Moreover, the entire road is blocked by CAV 2 and CAV 3 that are running at the same velocity shoulder-to-shoulder. Therefore, cooperative lane-changing is required for all vehicles to pass through the scenario efficiently.

The strategies generated by the proposed method are shown clearly in Fig. \ref{fig:overtaking}(c)-(f). Particularly in Fig. \ref{fig:overtaking}(c), it can be seen that when $t=1.5\,\textup{s}$, both CAV 3 and CAV 4 switch from lane 1 to lane 2 in order to clear the way for the fastest vehicle, namely CAV 1. After lane 1 is cleared, CAV 1 keeps proceeding forward to overtake all three CAVs that are currently in lane 2. After that, both CAV 2 and CAV 3 switch to lane 1 to overtake CAV 4, which is the slowest of all 4 CAVs. The solid dots in different colors demonstrate the historical location of the centers of all 4 CAVs and clearly show the lane-changing behaviors. During the entire overtaking process, the generated trajectories are collision-free. 

The velocities, accelerations, and steering angles for all 4 CAVs during the entire overtaking process are presented in Fig. \ref{fig:v-t-S}, with the colors of the curves corresponding to the colors of the vehicles. It can be seen from the curves of the velocities that originally CAV 2 and CAV 3 are proceeding at the same velocity, and then CAV 2 takes a brake for CAV 3 to cut in and yield the lane 1 free for CAV 1. After that, CAV 2 accelerates to bring its velocities back to the reference. The curves of acceleration and steering angle show that the control limits are satisfied.

\subsection{Roundabout}

\begin{figure}[t]
\centering
\includegraphics[scale=0.60]{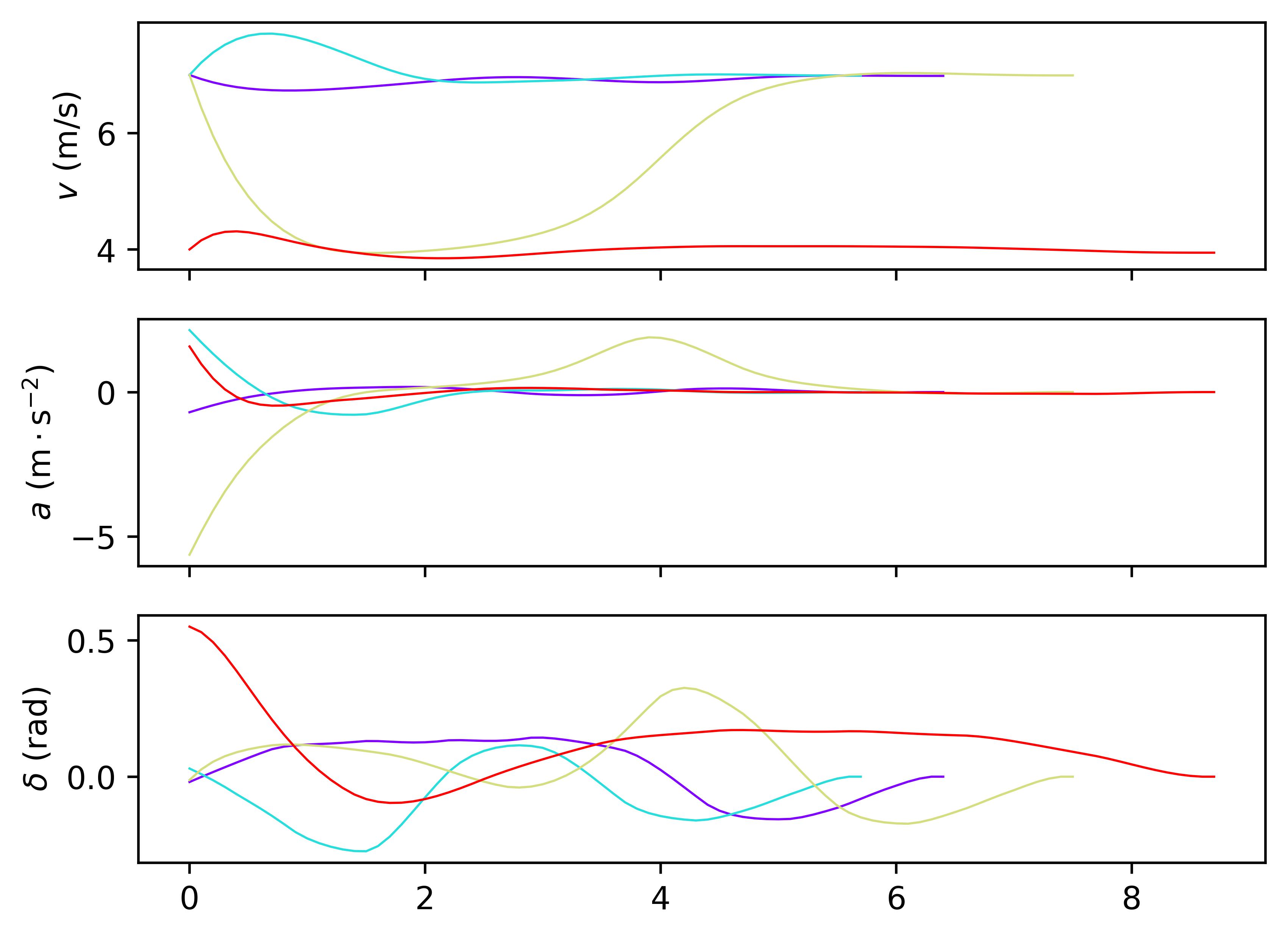}
\caption{Velocities, accelerations, and steering angles for all CAVs in the roundabout scenario.}
\label{fig:v-t-R}
\end{figure}

\begin{figure*}[t]

\subfigure[Waypoint graph]{\includegraphics[scale=0.53]{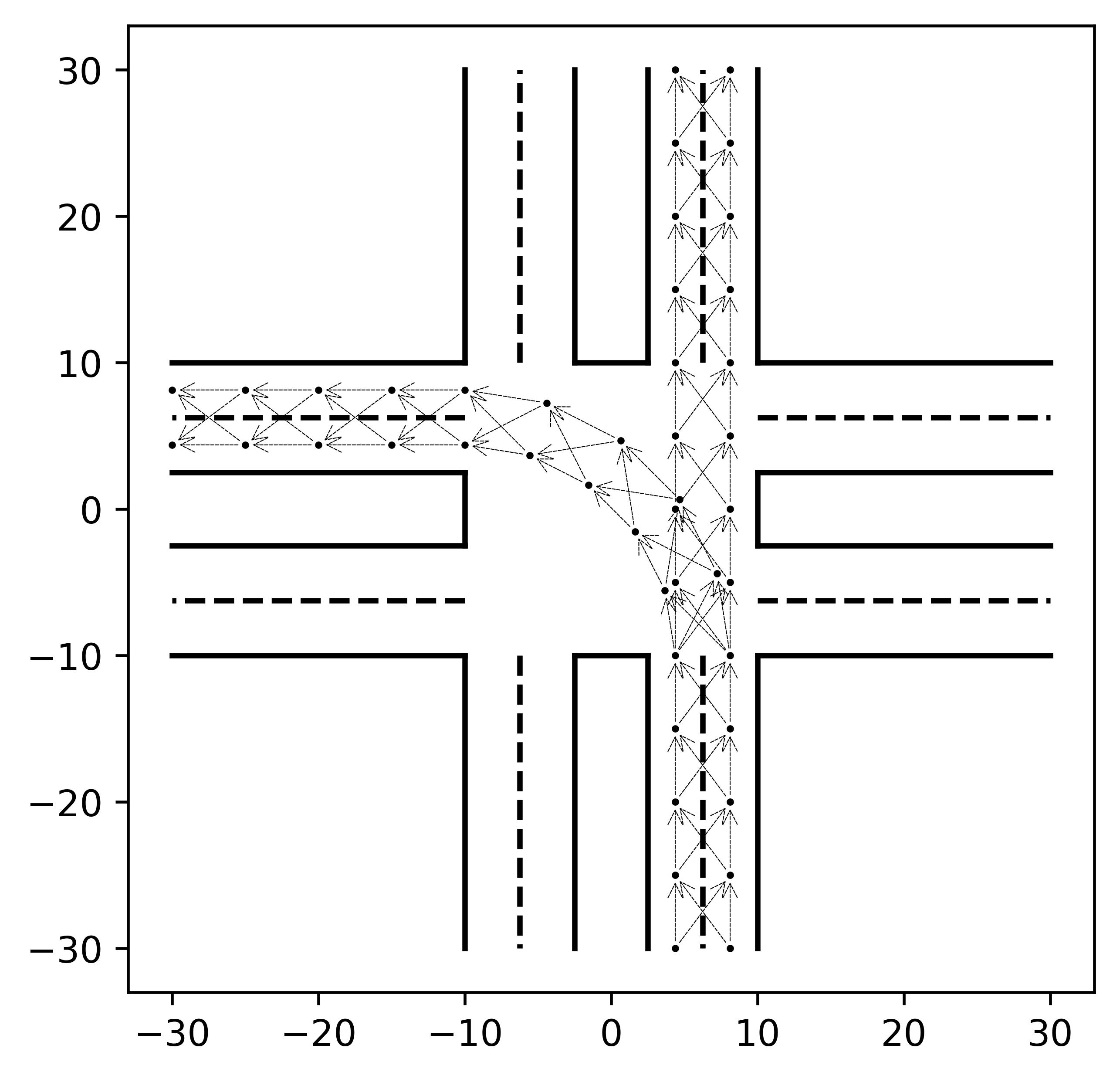}}
\centering
\subfigure[$t=0\,\textup{s}$]{\includegraphics[scale=0.53]{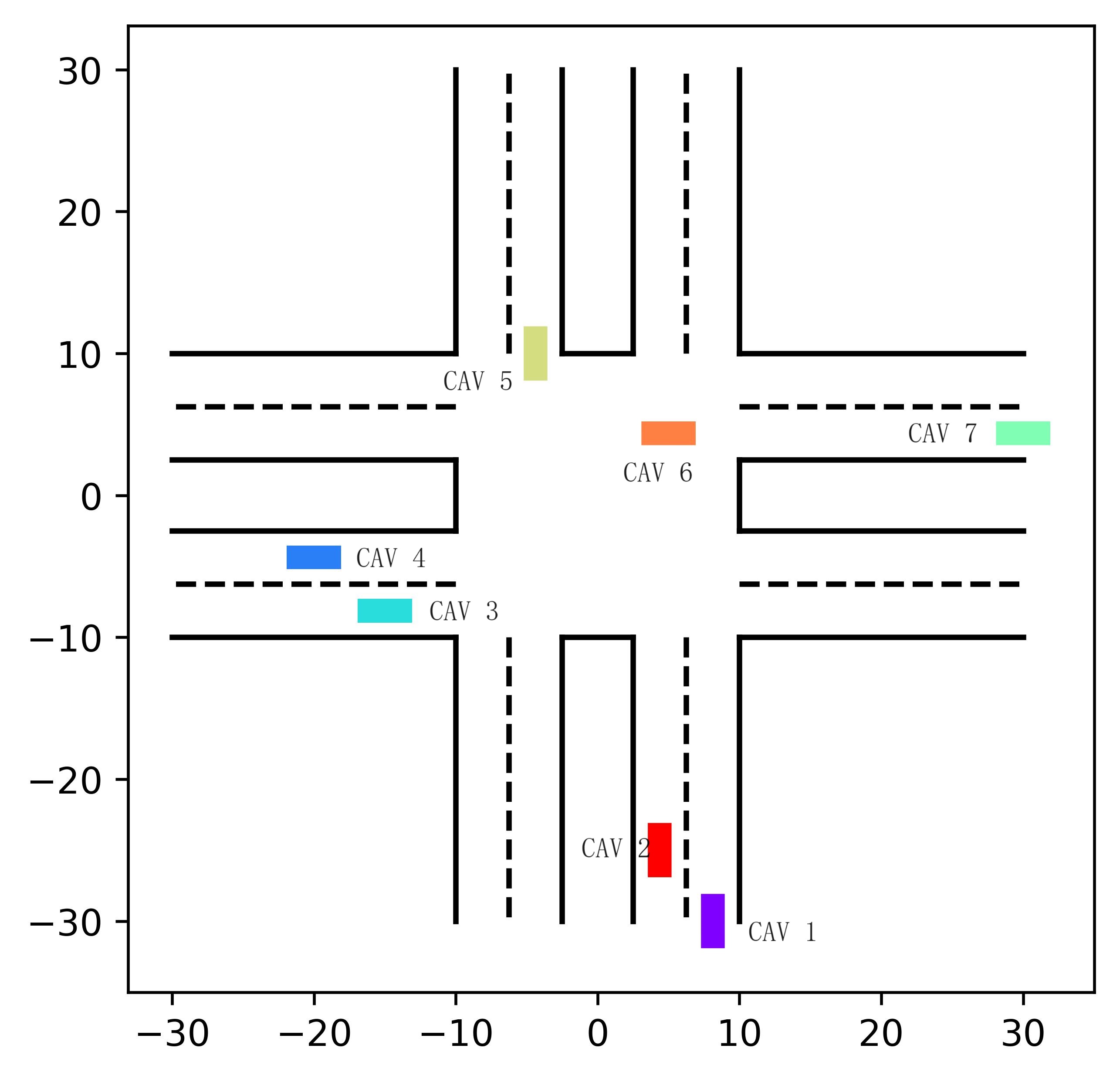}}
\subfigure[$t=0.6\,\textup{s}$]{\includegraphics[scale=0.53]{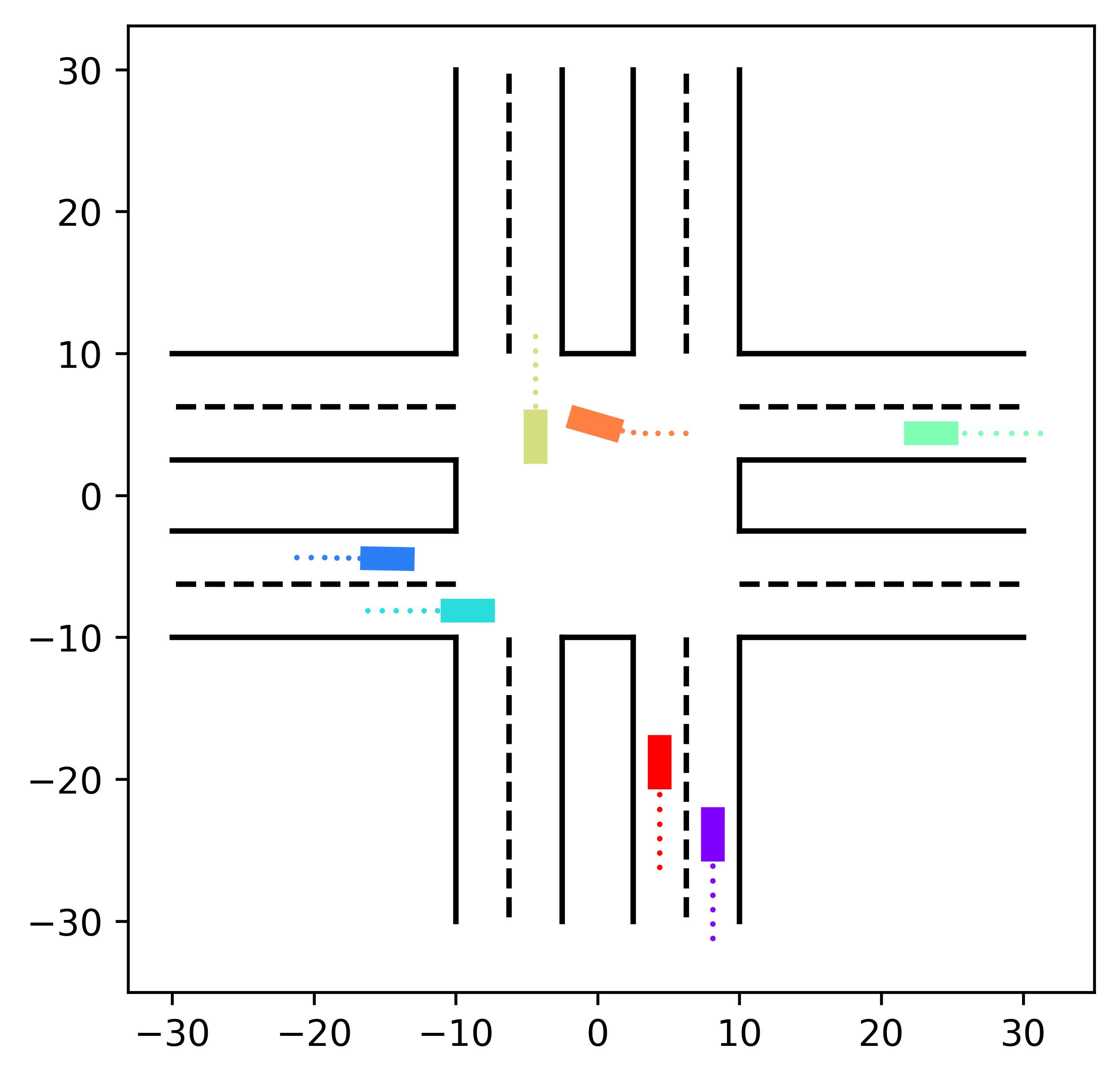}}
\subfigure[$t=2.3\,\textup{s}$]{\includegraphics[scale=0.53]{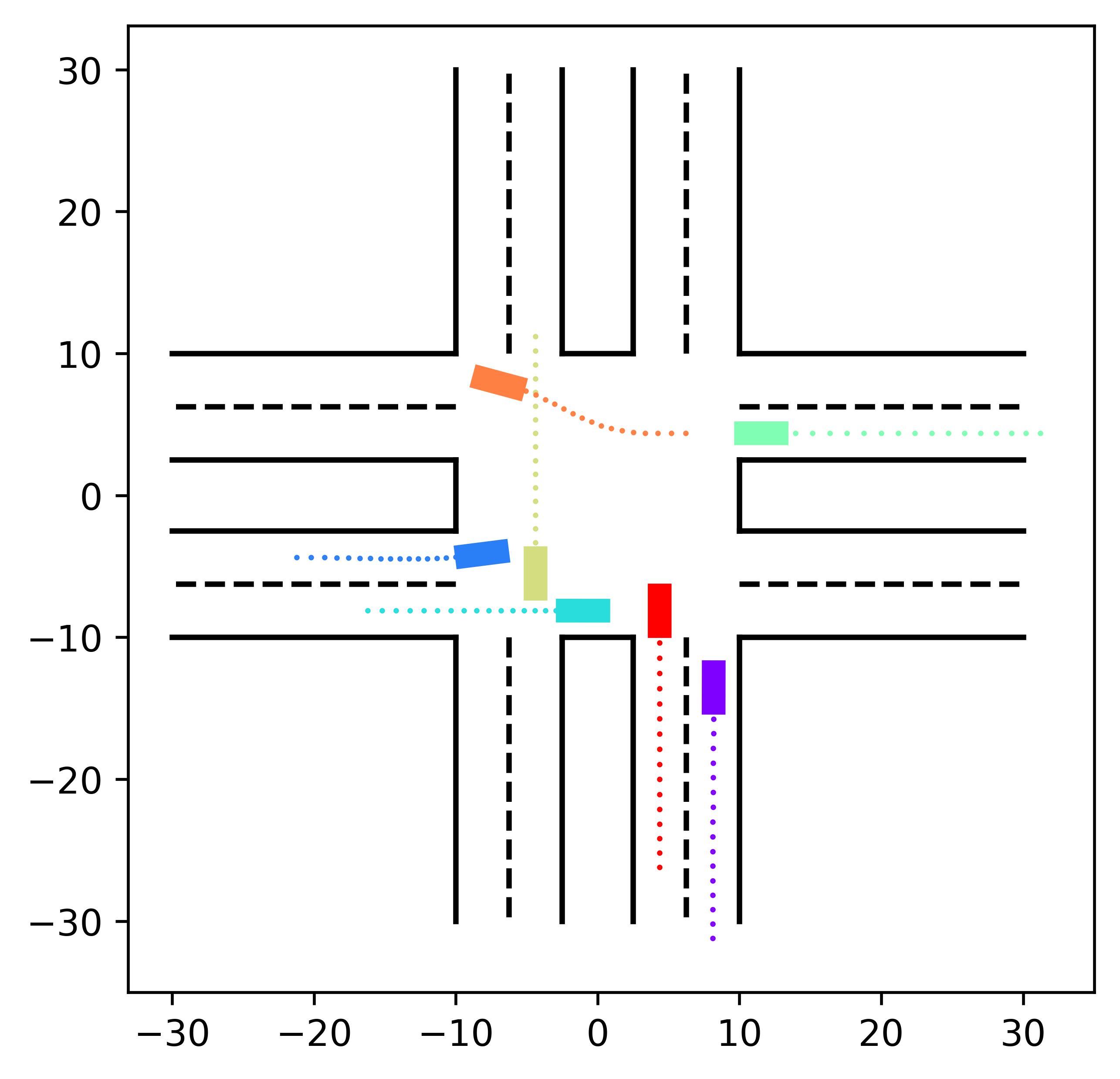}}
\subfigure[$t=3.4\,\textup{s}$]{\includegraphics[scale=0.53]{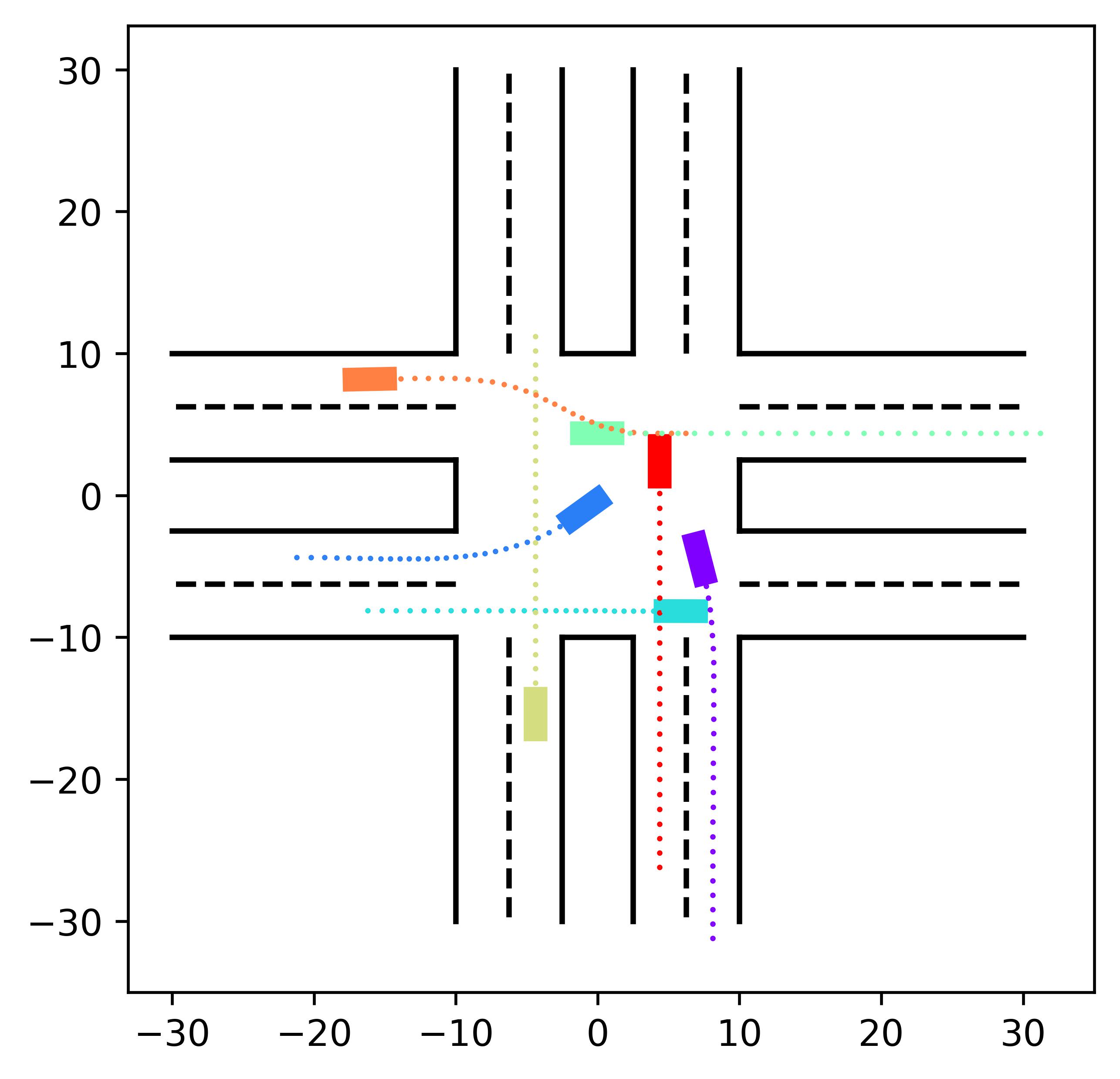}}
\subfigure[$t=4.5\,\textup{s}$]{\includegraphics[scale=0.53]{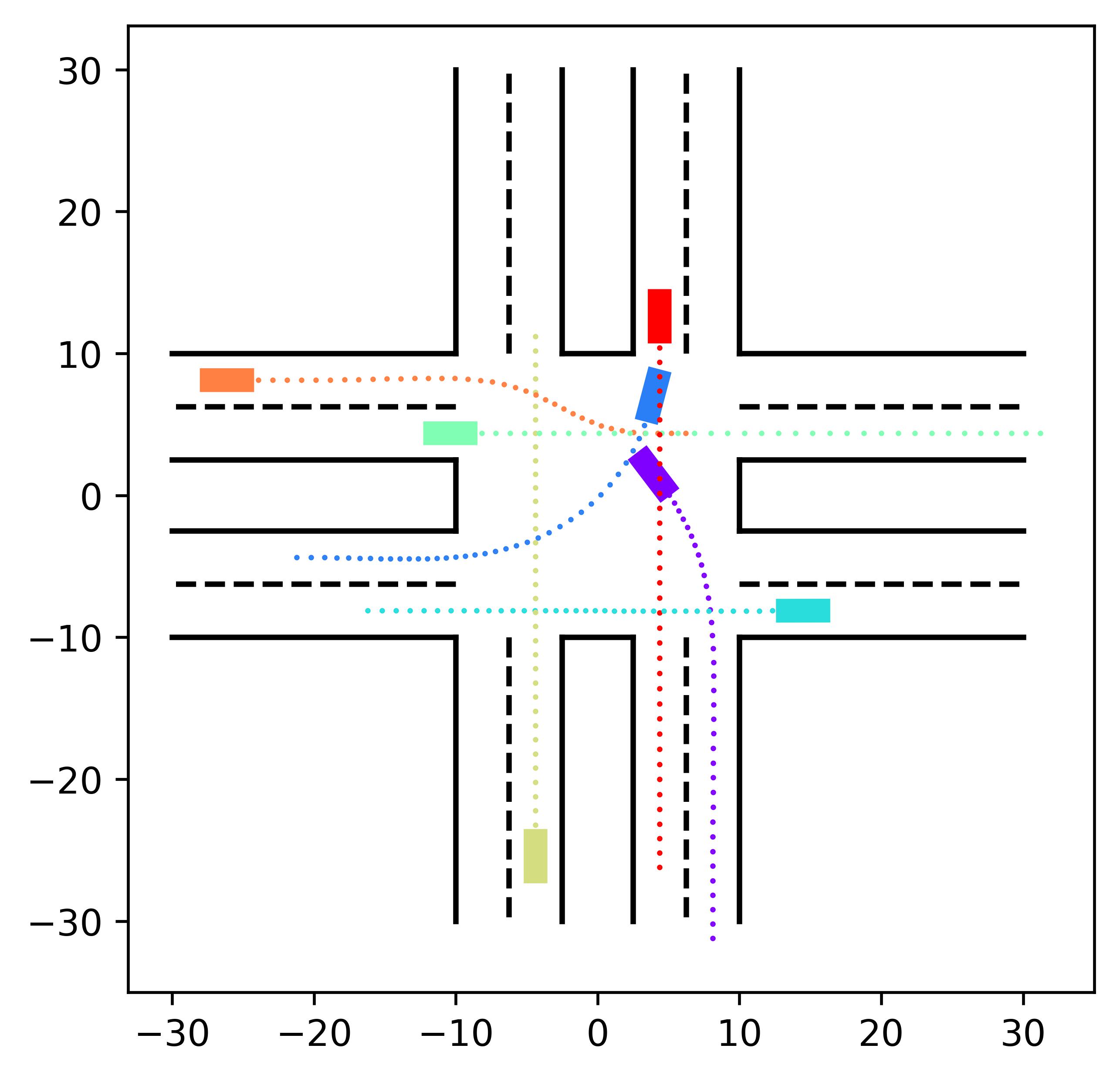}}

\caption{Simulation results for the unsignalized intersection scenario.}

\label{fig:cross}
\end{figure*}

% \begin{table}[t]
% \centering
% \caption{Intentions of CAVs in unsignalized roundabout}
% \begin{tabular}{p{0.5cm}p{1.3cm}p{0.5cm}p{1.3cm}p{0.5cm}p{1.0cm}}
% \hline
% No.               & Intention                            & No.              & Intention
% & No.              & Intention \\ \hline

% 1                      & left-turn                        & 2              & straight
% &3                & straight  \\

% 4       & left-turn &  5                & straight & 6                  & straight  \\

% 7         & straight   &    &      &  \\                     

% \hline
% \end{tabular}
% \end{table}

In this section, we present an example of cooperative decision-making in the roundabout scenario. In this example, behaviors including merging, overtaking, and exiting in the roundabout scenario are presented and explained. The constructed waypoint graph corresponding to part of the roundabout scenario is shown in Fig. \ref{fig:roundabout}(a). For simplicity and without loss of generality, the centerlines of lanes are assumed to be arcs, and waypoints are sampled along the centerlines with fixed interval arc angles. Each waypoint is connected to the next waypoint in the same lane. In particular, lane-switching is only allowed within the roundabout area and is prohibited at both the entrances and the exits. The initial poses of the 4 CAVs involved are shown in Fig. \ref{fig:roundabout}(b). The initial velocities for CAV 1-4 are set as 7\,m/s, 7\,m/s, 7\,m/s, and 4\,m/s, respectively, with reference velocities set to be the same as initial velocities. $V^i_{fast}$ and $V^i_{slow}$ are again set to be $1.3\,V^i_r$ and $0.6\,V^i_r$ respectively for all CAVs. In particular, CAV 1-3 aim to leave the roundabout from any of the exits below, while CAV 4 intends to keep proceeding inside the roundabout.

\begin{figure}[t]
\centering
\includegraphics[scale=0.60]{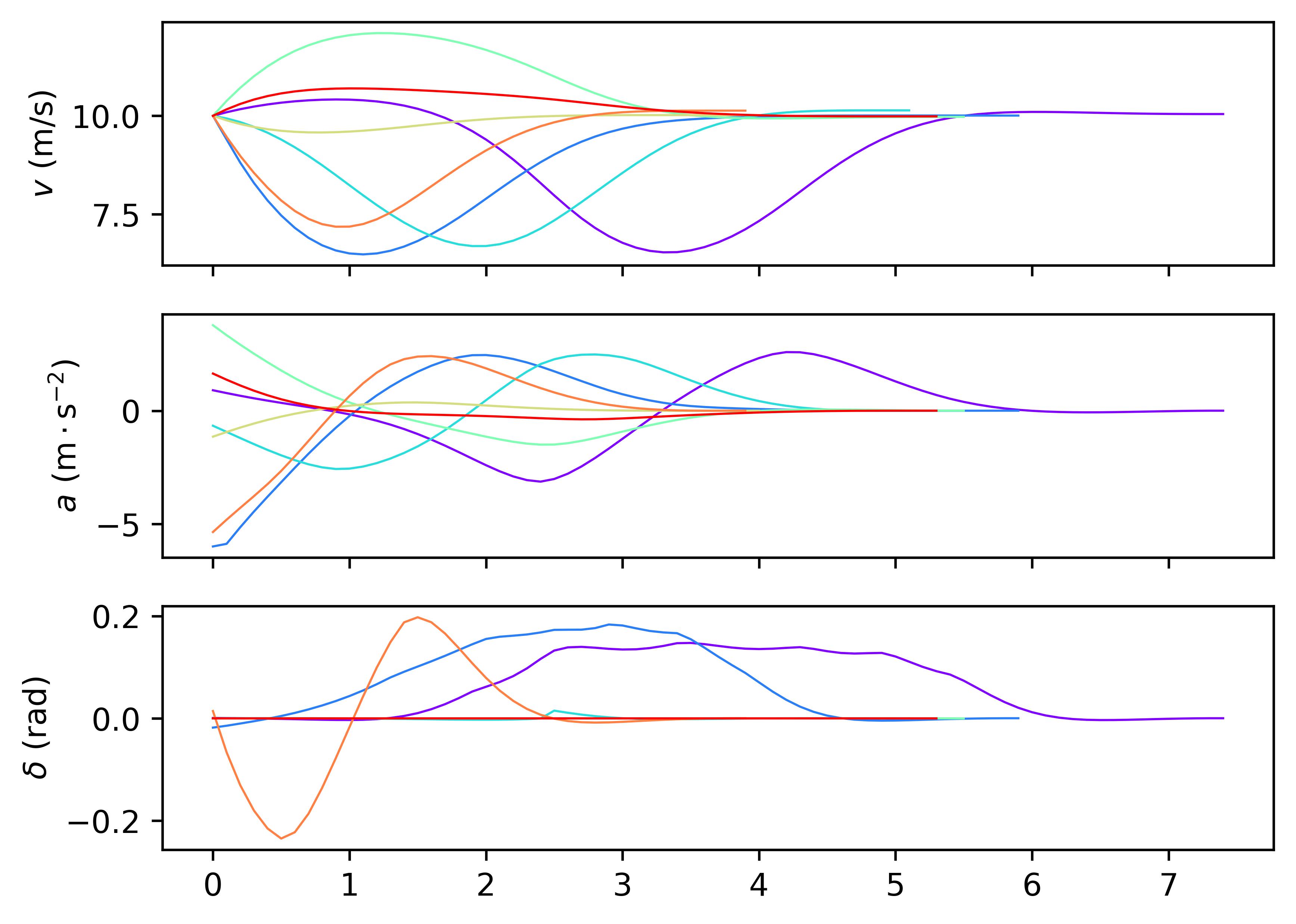}
\caption{Velocities, accelerations, and steering angles for all CAVs in the unsignalized intersection scenario.}
\label{fig:v-t-C}
\end{figure}

Fig. \ref{fig:roundabout}(c)-(h) illustrates the strategy generated. In particular, Fig. \ref{fig:roundabout}(c)(d) corresponds to the merging phase, and Fig. \ref{fig:roundabout}(e)-(f) corresponds to the overtaking and leaving phase. To facilitate the successful merging of CAV 3 into the roundabout area, CAV 4 switches into the inner lane to leave the critical merging zone clear, as is shown in Fig. \ref{fig:roundabout}(c). After that, CAV 3 successfully merges into the roundabout in front of CAV 1 (see Fig. \ref{fig:roundabout}(d)). Then both CAV 1 and CAV 3 overtake CAV 4 from the outer lane, with CAV 2 also switching to the outer lane and following (see Fig. \ref{fig:roundabout}(e)(f)). Eventually, CAV 1-3 leave the roundabout from the first exit below in a sequence, while CAV 4 keeps proceeding ahead (see Fig. \ref{fig:roundabout}(g)(h)). The trajectories are similarly demonstrated by the solid dots. No collision occurs during the entire process.

The velocities, accelerations, and steering angles for all 4 CAVs are shown in Fig. \ref{fig:v-t-R}. Some noteworthy behaviors demonstrated in the velocity curves include: 1) CAV 3 accelerates slightly in order to merge in front of CAV 1, and then returns to its reference velocity. 2) In the merging phase, CAV 2 takes a deep brake to avoid a rear-end collision with CAV 4. After both CAV 1 and CAV 3 overtake CAV 4 and leave the outer lane free for lane-switching, CAV 2 switches to the outer lane immediately and accelerates back to its reference velocity. Eventually, all the vehicles manage to maintain velocities close to their references. The acceleration and steering angle curves also demonstrate that the control limits are satisfied.

\subsection{Unsignalized Intersection}

We also perform simulations on an unsignalized intersection scenario to further verify the proposed method. In this example, only left-turn and straight-traveling are considered as most conflicts in unsignalized intersections are induced by these two behaviors. The geometry of the intersection and its corresponding waypoint graph are shown in Fig. \ref{fig:cross}(a). For brevity, we only show the waypoint graph corresponding to CAVs that enter the intersection from the lower-right entrance. Waypoints are sampled evenly along centerlines of straight lanes. The left-turn paths are represented by arcs, and waypoints are sampled along the paths with a fixed interval arc angle. Lane-changing is permitted in the entire scenario. The waypoint graphs that correspond to CAVs entering the intersection from other directions are similar rotated copies and therefore are omitted in Fig. \ref{fig:cross}(a). In this example, 7 CAVs traveling in different directions are involved. The initial poses of CAVs are shown in Fig. \ref{fig:cross}(b). In particular, CAV 1 and CAV 4 intend to turn left, and other CAVs intend to go straight. The initial velocities and reference velocities for all CAVs are 10\,m/s, and $V^i_{fast}$ and $V^i_{slow}$ are again set to be $1.3\,V^i_r$ and $0.6\,V^i_r$, respectively.

In the unsignalized intersection scenario, the paths of CAVs coming from different directions are usually intersecting with each other instead of overlapping. As a result, lane-switching is less relevant compared to the cases in the previous two scenarios. In general, it is easier for CAVs to resolve conflicts by simply accelerating and/or braking. In the simulation results, the only lane-switching behavior is performed by CAV 6. It switches to the left lane to avoid collision with CAV 5, which is a straight-traveling vehicle that cuts into its intended path from the lateral direction (see Fig. \ref{fig:cross}(c)(d)). Meanwhile, other CAVs avoid collisions by accelerating and decelerating in a fully cooperative manner to enable simultaneous crossing of the intersection in a compact and efficient manner (See Fig. \ref{fig:cross}(c)-(f)). The velocities, accelerations, and steering angles for CAVs are shown in Fig. \ref{fig:v-t-C}. The velocity curves demonstrate the accelerating and decelerating behaviors of each CAV to avoid colliding with each other, showing the effectiveness of the coordination strategy generated by the proposed method in resolving conflicts for the unsignalized intersection scenario. Again, the acceleration and steering angle curves show that the control limits are satisfied.

\subsection{Discussion}

%% In this section, three case studies are presented to demonstrate the application of the proposed method in traffic scenarios with substantially different geometries and topologies. Across these case studies, no adaptation is made to the proposed optimization scheme regarding the definition of optimization variables, construction of pertinent constraints, objective functions, and problem-solving methods. The only adaptation being made is in relation to the construction of the waypoint graph corresponding to each traffic scenario. Therefore, the proposed method is potentially generalizable to any traffic scenario as long as its topologies and geometries can be well represented by such a waypoint graph. Consequently, the cooperative decision-making problem of CAVs becomes readily solvable given the existence of a representative waypoint graph that reveals the topologies and geometries of the traffic scenario.

In this section, three case studies are presented to demonstrate the application of the proposed method in traffic scenarios with substantially different geometries and topologies. Across these case studies, no adaptation is made to the proposed optimization scheme regarding the definition of optimization variables, construction of pertinent constraints, objective functions, and problem-solving methods. The only adaptation being made is in relation to the construction of the waypoint graph corresponding to each traffic scenario. As such, our proposed method is potentially adaptable to any traffic scenario, provided its topologies and geometries can be accurately represented by a waypoint graph. This means that the cooperative decision-making problem for CAVs becomes readily solvable, given the existence of a representative waypoint graph that accurately reflects the topologies and geometries of the traffic scenario.

%% In the case studies presented above, waypoint graphs are constructed in a hand-crafted manner, which is laborious and time-consuming. The application of the introduced method to traffic scenarios with more complicated road structures and to a larger scale relies on High Definition Maps (HD Maps), which are considered critical infrastructure for autonomous driving. A typical HD Map contains important information regarding road structures, including accurate lane geometries and detailed lane topologies. These pieces of information are adequate for the construction of corresponding waypoint graphs. Nevertheless, extracting relevant information from HD Maps and constructing the corresponding waypoint graphs accordingly merely present engineering difficulties and thus are beyond the scope of this research.
In the case studies mentioned above, waypoint graphs are constructed in a hand-crafted manner, which is labor-intensive and time-consuming. The application of the introduced method to traffic scenarios with more complex road structures and on a larger scale relies on High Definition Maps (HD Maps), which are deemed critical infrastructure for autonomous driving. A typical HD Map contains vital information about road structures, including accurate lane geometries and detailed lane topologies. This information is sufficient for the construction of corresponding waypoint graphs. However, extracting the relevant information from HD Maps and constructing the corresponding waypoint graphs are engineering challenges that are beyond the scope of this research.

\section{Conclusion}
%% In this paper, a generic optimization scheme is introduced, which aims to tackle the cooperative decision-making problem of CAVs across a variety of urban traffic scenarios given arbitrary road topologies. By formulating the problem as an MILP, paths and respective time profiles for all CAVs are optimized jointly. With nonlinear constraints relaxed, solutions are readily available and optimalities are attainable by using common off-the-shelf solvers. To verify the effectiveness of the proposed method, simulations on three traffic scenarios with substantially different geometries and topologies are performed. Based on the simulation results, the developed optimization scheme is able to generate reasonable strategies for CAVs in each of the scenarios without adaptations, and thus the generalization ability of the proposed approach is demonstrated. 

In this paper, a generic optimization scheme is introduced, which aims to tackle the cooperative decision-making problem of CAVs across a variety of urban traffic scenarios given generic road topologies. By formulating the problem as an MILP, paths and respective time profiles for all CAVs are optimized jointly.
By relaxing nonlinear constraints, solutions are readily attainable and optimalities can be achieved using numerical solvers. To verify the effectiveness of the proposed method, simulations were conducted on three traffic scenarios with significantly different geometries and topologies. The simulation results show that the developed optimization scheme can generate reasonable strategies for CAVs in each scenario without  adaptations, thereby demonstrating the generalization capability of the proposed approach.
%% Possible future works include introducing rule-based heuristics to enhance the overall computational efficiency. Meanwhile, since the overall cost minimized in the MILP is essentially a potential function, the entire problem can be reformulated as a potential game to model and cope with the self-interest-oriented behaviors of traffic participants.
Potential future works involve the incoporation of rule-based heuristics to improve the overall computational efficiency. Moreover, since the overall cost minimized in the MILP essentially serves as a potential function, the entire problem could be reformulated as a potential game. This would model and manage the self-interest-oriented behaviors of traffic participants appropriately.

\bibliographystyle{IEEEtran}
\bibliography{refs}

\end{document}